\documentclass[runningheads]{llncs}
\usepackage{graphicx}
\usepackage{amsmath,amssymb} 
\usepackage{color}

\usepackage{epsfig}
\usepackage{comment}
\usepackage{booktabs}
\usepackage{dirtytalk}
\usepackage{amssymb}
\usepackage{mathtools}
\usepackage{sidecap}
\usepackage[normalem]{ulem}
\usepackage{hyperref}  
\usepackage{multirow}
\usepackage{subfigure}

\usepackage[usenames,dvipsnames,svgnames,table]{xcolor}
\usepackage{xspace}
\makeatletter
\DeclareRobustCommand\onedot{\futurelet\@let@token\@onedot}
\def\@onedot{\ifx\@let@token.\else.\null\fi\xspace}
\def\eg{\emph{e.g}\onedot} 

\def\ie{\emph{i.e}\onedot}

\def\etal{\emph{et al}\onedot}
\makeatother

\def\Vec#1{{\boldsymbol{#1}}}
\def\Mat#1{{\boldsymbol{#1}}}

\DeclareMathOperator*{\argmin}{arg\,min}

\usepackage{float}

\begin{document}
\pagestyle{headings}
\mainmatter

\def\ACCV20SubNumber{334}  

\title{Channel Recurrent Attention Networks for Video Pedestrian Retrieval} 
\titlerunning{Channel Recurrent Attention Networks}
%
\author{Pengfei Fang~\inst{1,2}\orcidID{0000-0001-8939-0460} \and
Pan Ji\thanks{Work done while at NEC Laboratories America}~\inst{3}\orcidID{0000-0001-6213-554X} \and
Jieming Zhou~\inst{1,2}\orcidID{0000-0002-3880-6160} \and
Lars Petersson~\inst{2}\orcidID{0000-0002-0103-1904} \and
Mehrtash Harandi~\inst{4}\orcidID{0000-0002-6937-6300}}
\authorrunning{P. Fang et al.}
%
\institute{The Australian National University \\
\email{Pengfei.Fang@anu.edu.au}\and DATA61-CSIRO \and OPPO US Research Center \and Monash University}

\maketitle

\setcounter{footnote}{0}

\begin{abstract}
Full attention, which generates an attention value per element of the input feature maps, has been successfully demonstrated to be beneficial in visual tasks. In this work, we propose a fully attentional network, termed {\it channel recurrent attention network}, for the task of video pedestrian retrieval. The main attention unit, \textit{channel recurrent attention}, identifies attention maps at the frame level by jointly leveraging spatial and channel patterns via a recurrent neural network. This channel recurrent attention is designed to build a global receptive field by recurrently receiving and learning the spatial vectors. Then, a \textit{set aggregation} cell is employed to generate a compact video representation. Empirical experimental results demonstrate the superior performance of the proposed deep network, outperforming current state-of-the-art results across standard video person retrieval benchmarks, and a thorough ablation study shows the effectiveness of the proposed units. 
\end{abstract}

\section{Introduction}\label{sec:intro}
This work proposes \textit{Channel Recurrent Attention Networks} for the purpose of pedestrian retrieval\footnote{For the remainder of this paper, we shall use the terms ``pedestrian retrieval'' and ``person re-identification'' interchangeably.}, in  challenging video data. 

Pedestrian retrieval, or person re-identification (re-ID), a core task when tracking people across camera networks~\cite{ZhengLiang2016arXivReIDPastPresentFuture}, attempts to retrieve all correct matches of a person from an existing database, given a target query. 
There are many challenges to this task, with a majority stemming from a poor quality or large variation of the captured images. This often leads to difficulties in building a discriminative representation, which in turn results in a retrieval system to mismatch its queries. Video-, as opposed to single image-, person re-ID offers the possibility of a richer and more robust representation as \textcolor{black}{temporal} cues can be utilized to obtain a compact, discriminative and robust video representation for the re-ID task. In many practical situations, the retrieval performance suffers from spatial misalignment~\cite{Suh_2018_ECCV,LiWei2018CVPRHarmoniousAttentionNetforReID,Jim_ivc}, caused by the movement of body parts, which affects the retrieval machine negatively. Focusing on this issue, many efforts have been made to develop visual attention mechanisms~\cite{LiWei2018CVPRHarmoniousAttentionNetforReID,Mancs,Pengfei_2019_ICCV,Chen_2019_ICCV_MHO,COSAM_2019_ICCV}, which makes the network attend to the discriminative areas within person bounding boxes, relaxing the constraints stemming from spatial nuances.

Attention mechanisms have been demonstrated to be successful in various visual tasks, such as image classification~\cite{Hu_2018_CVPR,Woo_2018_ECCV_CBAM}, object detection~\cite{WangXiaolong2017CVPRNonLocalNN}, scene segmentation~\cite{Fu_2019_CVPR_DualAtt,li2018deep} to name just a few. Generally speaking, attention mechanisms can be grouped into channel attention~\cite{Hu_2018_CVPR}, spatial attention~\cite{Wang_2017_CVPR_RAN}, and full attention~\cite{Mancs}, according to the dimensions of the generated attention maps. The channel attention usually summarizes the global spatial representation of the input feature maps, and learns a channel pattern that re-weights each slice of the feature maps. In contrast, the spatial attention learns the spatial relationships within the input feature maps and re-weights each spatial location of the feature maps. Lastly, full attention not only learns the channel patterns, but also preserves spatial information in the feature maps, which significantly improves the representation learning~\cite{hjelm2018learning}.

Various types of full attention mechanisms have been studied extensively for the task of pedestrian retrieval~\cite{Mancs,Pengfei_2019_ICCV,Chen_2019_ICCV_MHO}. In~\cite{Mancs}, the fully attentional block re-calibrates the channel patterns by a non-linear transformation. Thereafter, higher order channel patterns are exploited to attend to the channel features~\cite{Pengfei_2019_ICCV,Chen_2019_ICCV_MHO}. However, the aforementioned attention fails to build {\it long-range} spatial relationships due to the use of a $1\times 1$ convolution. The work in~\cite{LiWei2018CVPRHarmoniousAttentionNetforReID} learns spatial interactions via a convolutional layer with a larger kernel size ($3\times 3$), but the attention module therein still only has a small spatial receptive field. In visual attention, we want the network to have the capacity to view the feature maps globally and decide what to focus on for further processing~\cite{WangXiaolong2017CVPRNonLocalNN}. A global view can be achieved by applying fully connected (FC) layers, which, unfortunately, introduces a huge number of learnable parameters if implemented naively.

In this work, we propose a full attention mechanism, termed {\it channel recurrent attention}, to boost the video pedestrian retrieval performance. The channel recurrent attention module aims at creating a global view of the input feature maps. Here, the channel recurrent attention module benefits from the recurrent operation and the FC layer in the recurrent neural network. We feed the vectorized spatial map
to the Long Short Term Memory (LSTM) sequentially, such that the recurrent operation of the LSTM captures channel patterns while the FC layer in the LSTM has a global receptive field of each spatial slice. To handle video data, we continue to develop a \textit{set aggregation} cell, which aggregates the frame features into a discriminative clip representation. In the set aggregation cell, we re-weight each element of the corresponding frame features, in order to selectively emphasize useful features and suppress less informative features, with the aid of the associated clip features. The \textbf{contributions} of this work include:
\begin{itemize}
\item[$\bullet$] The proposal of a novel channel recurrent attention module to jointly learn spatial and channel patterns of each frame feature map, capturing the global view of the feature maps. To the best of the authors' knowledge, this is the first attempt to consider the global spatial and channel information of feature maps in a full attention design for video person re-ID.

\item[$\bullet$] The development of a simple yet effective set aggregation cell, which aggregates a set of frame features into a discriminative clip representation.

\item[$\bullet$] State-of-the-art performance across standard video re-ID benchmarks by the proposed network. The generalization of the attention module is also verified by the competitive performance on the single image re-ID task.
\end{itemize}

\section{Related Work} \label{sec:relatedwork}

This section summarizes the related work of pedestrian retrieval and relevant attention mechanisms.

Several approaches have been investigated to improve the state-of-the-art retrieval performance for both single image and video person re-ID~\cite{ZhengLiang2016arXivReIDPastPresentFuture,GongShaogang2014SpringerPersonReID}. Focusing on deep neural networks~\cite{lecun2015deep}, metric learning and representation learning are the two dominating approaches in modern person re-ID solutions~\cite{ZhengLiang2016arXivReIDPastPresentFuture}. In \cite{YiDong2014ICPRDeepMetricLearningforReID}, the similarity of input pairs is calculated by a siamese architecture~\cite{BromleyNIPS1994Siamese}. The improved deep siamese network learns an image-difference metric space by computing the cross-input relationships~\cite{AhmedEjaz2015CVPR}.

In representation learning, the pedestrian is represented by the concatenation of multi-level features along the deep network ~\cite{WuShuanxuan2016WACVAnenhancedDeepfeatureRepresentationforReID} or by combining the person appearance and body part features~\cite{LiWei2018CVPRHarmoniousAttentionNetforReID}.    
Beyond single image-based person re-ID methods, efficient temporal modeling~\cite{gao2018revisiting} is further required when working with video clips. This is challenging as there is a need to create a compact video representation for each identity. McLaughlin~\etal proposed average and max temporal pooling for aggregating frame features and each frame feature is the output of a recurrent neural network~\cite{McLaughlin2016CVPRRNNforVideoReID}.

Recent work has shown that person re-ID benefits significantly from attention mechanisms highlighting the discriminative areas inside the person bounding boxes when learning an embedding space~\cite{LiuHao2017arXivEndtoEndComparativeAttentionNetworksforReID,Liu_2017_ICCV,LiWei2018CVPRHarmoniousAttentionNetforReID,Mancs,Pengfei_2019_ICCV,Chen_2019_ICCV_MHO}. In~\cite{LiuHao2017arXivEndtoEndComparativeAttentionNetworksforReID,Liu_2017_ICCV}, the spatial attention mask is designed to attend one target feature map or various feature maps along the deep network. In~\cite{Mancs}, a fully attentional block is developed to re-calibrate the channel features. Second or higher order statistical information is also employed in full attention frameworks~\cite{Pengfei_2019_ICCV,Chen_2019_ICCV_MHO}. The full attention shape map is also generated in the harmonious attention module~\cite{LiWei2018CVPRHarmoniousAttentionNetforReID}, by integrating channel attention and spatial attention. The aforementioned attention mechanism either fails to build spatial-wise relationships, or receives a limited spatial receptive field. Unlike the above methodology of full attention, we intend to develop an attention mechanism which preserves the advantage of the common full attention, while also perceiving a global spatial receptive field of the feature maps.

\section{Method} \label{sec:method}
This section details the proposed deep network in a top-down fashion: starting with the problem formulation of the application, followed by the network architecture and the main attention module in the network, namely, the channel recurrent attention module. Thereafter, we also introduce a set aggregation cell, to encode a compact clip representation.

\paragraph{\bf{Notation.}}
We use $\mathbb{R}^n$, $\mathbb{R}^{h \times w}$, $\mathbb{R}^{c \times h \times w}$ and $\mathbb{R}^{t \times c \times h \times w}$ to denote the $n$-dimensional Euclidean space,  the real matrix space (of size ${h \times w}$), and the image and video spaces, respectively. A matrix or vector transpose is denoted by the superscript $\top$. The symbol $\odot$ and $\oplus$, represent the Hadamard product (\ie, element-wise multiplication) and element-wise summation. ${\sigma:\mathbb{R} \to [0,1]}$ is the $\mathrm{sigmoid}$ function. $\mathrm{BN}: \mathbb{R}^n \to \mathbb{R}^n, \mathrm{BN}(\Vec{x}) \coloneqq \gamma \frac{\Mat{x} - \mathrm{E}[\Mat{x}]}{\sqrt{\mathrm{Var}[\Mat{x}]}} + \beta$ and $\mathrm{ReLU}: \mathbb{R} \to \mathbb{R}_{\geq 0}, \mathrm{ReLU}(x) \coloneqq \mathrm{max}(0, x)$ refer to batch normalization and rectified linear unit. $\phi$, $\varphi$, $\varpi$ and $\psi$ are used to represent embedding functions (\eg, linear transformations, or self-gating layer).

\subsection{Problem Formulation}
Let a fourth-order tensor, $\Mat{T}_i = [T_i^1, T_i^2, \ldots, T_i^N] \in \mathbb{R}^{N \times C \times H \times W}$, denote the $i$-th video sequence of a pedestrian, where $N$, $C$, $H$, and $W$ are the number of frames, channels, height and width, respectively. Each video sequence $\Mat{T}_i$ is labeled by its identity, denoted by $y_i \in \{1, \ldots, k\}$. The training set with $M$ video sequences is described by $\mathbb{T} = \{\Mat{T}_i, y_i\}_{i = 1}^M$. The video person re-ID model, $f_{\theta}:\mathcal{T} \to \mathcal{F}$, describes a non-linear embedding from the video space, $\mathcal{T}$, to an embedding space, $\mathcal{F}$, in which the intra-class/person distance is minimized and the inter-class/person distance is maximized. The target of training a deep neural network is to learn a set of parameters, $\theta^\star$, with minimum loss value (\eg, $\mathcal{L}$), satisfying: $\theta^\star = \argmin_{\theta}\sum_{i= 1}^M{\mathcal{L}(f_{\theta}(\Mat{T}_i), y_i)}$. {In the training stage, we randomly sample batches of video clips, where each video clip has only $t$ frames (randomly chosen). Such frames are order-less and hence, we are interested in set-matching for video re-ID.}

\subsection{Overview}
\begin{figure}[ht]
\centering
\includegraphics[width = \textwidth]{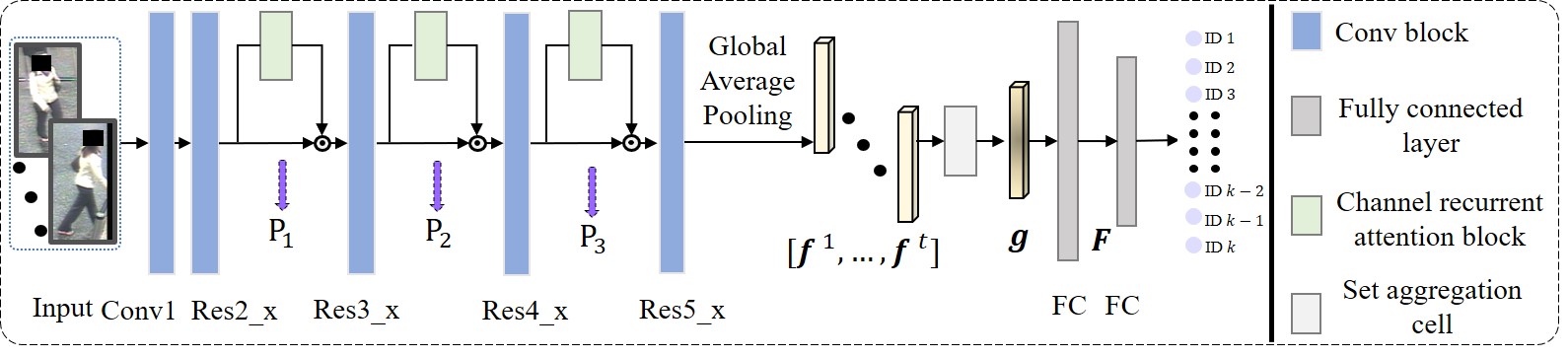}

\caption{\footnotesize{The proposed deep neural network with channel recurrent attention modules and a set aggregation cell.}}\label{fig:Strucutre}
\end{figure}

We begin by providing a sketch of our design first. In video person re-ID, one would ideally like to make use of a deep network to extract the features of the frames and fuse them into a compact and discriminative clip-level representation. In the lower layers of our design, we have five convolutional blocks  along with channel recurrent attention modules at positions  $\text{P}_1$,  $\text{P}_2$ and $\text{P}_3$ (see Fig.~\ref{fig:Strucutre}). Once the deep network extracts a set of frame features (\ie, $[\Vec{f}^1, \ldots, \Vec{f}^t]$ in Fig.~\ref{fig:Strucutre}), a set aggregation cell is utilised to fuse frame features into a compact clip-level feature representation (\ie, $\Vec{g}$). The final clip representation is $\Vec{F} = \mathrm{ReLU}\big( \mathrm{BN}(\Mat{W}_1^{\top} \Vec{g}) \big)$, followed by another FC layer to perform identity prediction (\ie, $p = \Mat{W}_2^{\top}\Vec{F}$), where $\Mat{W}_1, \Mat{W}_2$ are the learnable parameters in the FC layers. We note that the output of the middle convolutional layers captures rich spatial and channel information~\cite{WangXiaolong2017CVPRNonLocalNN,Pengfei_2019_ICCV}, such that the attention modules can make better use of this available information.

The network training benefits from multi-task learning, which formulates the network training as several sub-tasks. Our work follows~\cite{gao2018revisiting}, and trains the network using a triplet loss and a cross-entropy loss. The details of the loss functions are described in the supplementary material.

\subsection{Channel Recurrent Attention}

\begin{figure}[ht] 
\centering
\includegraphics[width = 0.9\linewidth]{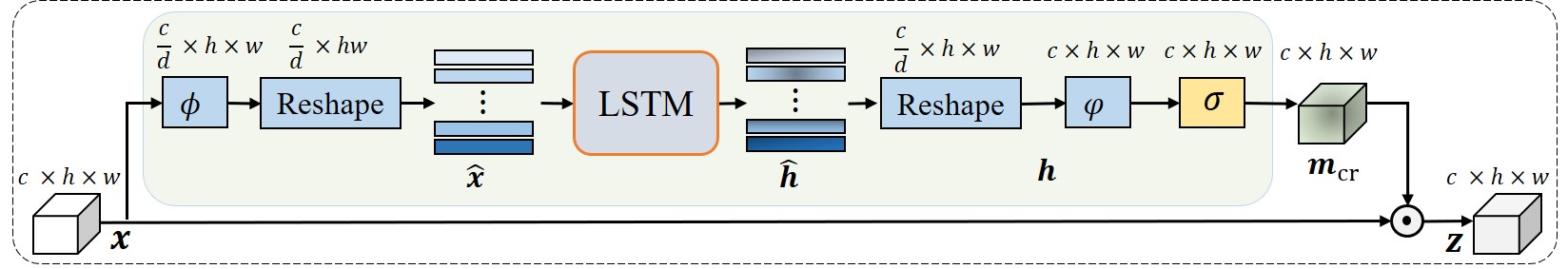}
\caption{\footnotesize{The architecture of the proposed channel recurrent attention module.}} \label{fig:RA}
\end{figure}

We propose the channel recurrent attention module (see Fig.~\ref{fig:RA}), which learns the spatial and channel patterns globally in a collaborative manner with the assistance of an LSTM, over the feature maps of each frame. To be specific, we model the input feature maps as a sequence of spatial feature vectors, and feed it to an LSTM to capture global channel patterns by its recurrent operation. In our design, the hidden layer (\eg, FC) of the LSTM unit, can be understood as having a global receptive field, acting on each spatial vector while sharing weights with other spatial vectors, addressing the limitation of a small receptive field in CNNs. In~\textsection\ref{sec:expt}, our claim is empirically evaluated in an ablation study.

Let ${\Mat{x}} \in \mathbb{R}^{{c} \times h \times w}$ be the input of the channel recurrent attention module. In our implementation, we project ${{\Mat{x}}}$ to $\phi({{\Mat{x}}})$, reducing the channel dimension by a ratio of $1/d$, and reshape the embedded tensor $\phi({{\Mat{x}}})$ to a matrix $\hat{\Mat{x}} = [\hat{x}_1, \ldots, \hat{x}_{\frac{{c}}{d}}]^{\top} \in \mathbb{R}^{\frac{{c}}{d} \times hw}$, where a row of $\hat{\Mat{x}}$ (\eg, $\hat{x}_i \in \mathbb{R}^{hw}, i = 1, \ldots, \frac{{c}}{d}$) denotes the spatial vector of a slice. {The effect of the ratio $1/d$ is studied in \textsection\ref{ablationstudy}.} A sequence of spatial vectors is then fed to an LSTM unit and the LSTM generates a sequence of hidden states, in matrix form:

\begin{equation}
\begin{split}
\hat{\Mat{h}}  &= \mathrm{LSTM}(\hat{\Mat{x}}) = [\hat{h}_1, \ldots, \hat{h}_{\frac{{c}}{d}}]^{\top},
\end{split}
\end{equation}
where $\hat{h}_i \in \mathbb{R}^{hw},  i = 1, \ldots, {c}/d$ is a sequence of hidden states and $\mathrm{LSTM}(\cdot)$ represents the recurrent operation in an LSTM. The insight is illustrated by the unrolled LSTM, shown in Fig.~\ref{cra_1}. $\hat{\Mat{h}}$ is further reshaped to the same size as the input tensor $\phi({\Mat{x}})$ (\ie, $\Mat{h} = \mathrm{Reshape}(\hat{\Mat{h}}), \Mat{h}\in \mathbb{R}^{\frac{{c}}{d} \times h \times w}$). The final attention value is obtained by normalizing the embedded $\Mat{h}$, written as:

\begin{equation}
\begin{split}
\Mat{m}_{\mathrm{cr}} = {\sigma}(\varphi(\Mat{h})).
\end{split}
\end{equation}
Here, $\varphi(\Mat{h}), \Mat{m}_{\mathrm{cr}} \in \mathbb{R}^{c \times h \times w}$. This normalized tensor acts as a full attention map and re-weighs the elements of the associated frame feature map (see Fig.~\ref{fig:RA}), by element-wise multiplication:
\begin{equation}\label{eq:zz}
\Mat{z} = \Mat{m}_{\mathrm{cr}} \odot \Mat{x}.
\end{equation}

\begin{figure}[H]
\centering
	\subfigure[Feature aggregation~\cite{DeepPerson_PR}]{\includegraphics[width=2.4cm,height=1cm]{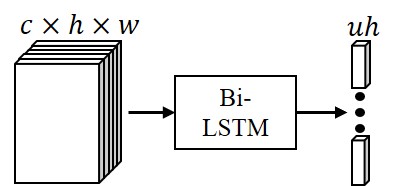}\label{1}}%
	\quad
	\subfigure[Feature aggregation~\cite{RFA-Net}]{\includegraphics[width=2.2cm,height=0.8cm]{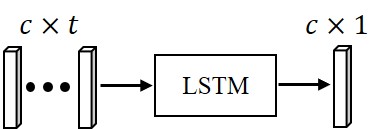}\label{2}}%
	\quad
	\subfigure[Attention mask generation~\cite{LiuHao2017arXivEndtoEndComparativeAttentionNetworksforReID,DVAN}]{\includegraphics[width=3cm,height=1.15cm]{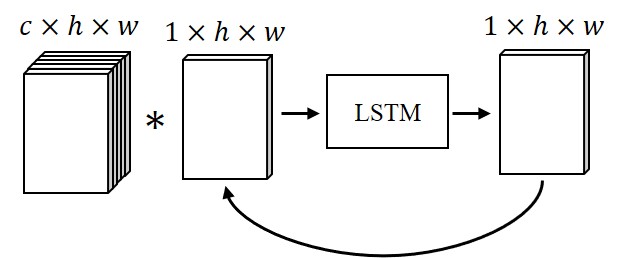}\label{3}}%
	\quad
	\subfigure[Attention maps generation (Ours)]{\includegraphics[width=3.1cm,height=1.4cm]{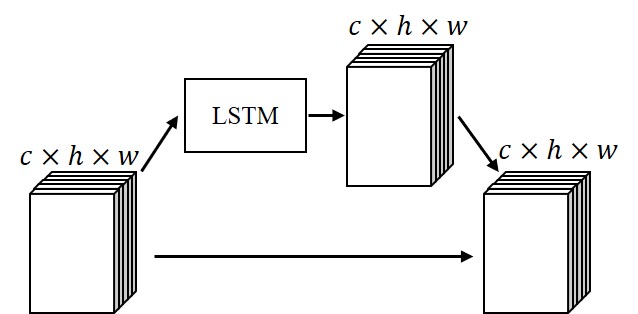}\label{4}}%
	\quad
	\caption{{Schematic comparison of our attention mechanism and existing LSTM-based works. In (c), the notation $*$ denotes a weighted sum operation.}}\label{fig:diff}
\end{figure}

\textcolor{black}{
\begin{remark}
There are several studies that use LSTMs to aggregate features~\cite{DeepPerson_PR,RFA-Net} (see Fig.~\ref{1} and \ref{2}), or generate attention masks~\cite{LiuHao2017arXivEndtoEndComparativeAttentionNetworksforReID,DVAN} (see Fig.~\ref{3}). Our channel recurrent attention module (see Fig. \ref{4}) is significantly different from existing works as shown in Fig.~\ref{fig:diff}. The designs in \cite{DeepPerson_PR} and \cite{RFA-Net} employ an LSTM to  {\it aggregate} features either from input feature maps~\cite{DeepPerson_PR}, or a sequence of frame features in a video~\cite{RFA-Net}. In~\cite{LiuHao2017arXivEndtoEndComparativeAttentionNetworksforReID,DVAN}, an attention value for each spatial position of the feature maps (\ie, spatial attention) is constructed recursively, while ignoring the relation in the channel dimension. In contrast, our channel recurrent attention generates an attention value per element of the feature maps (\ie, full attention), thereby enabling the ability to learn richer spatial and channel features.   
\end{remark}
}

\subsection{Set Aggregation}
\begin{figure}[ht] 
\centering
\includegraphics[width = 0.85\linewidth]{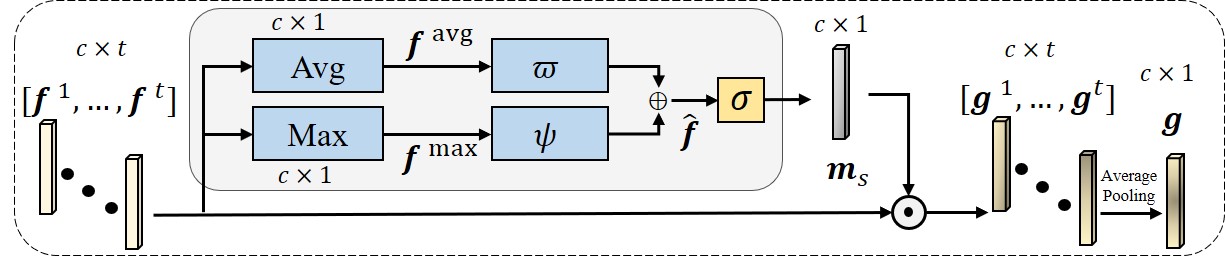}
\caption{\footnotesize{The architecture of the proposed set aggregation cell.}} \label{fig:SA}
\end{figure}

To encode a compact clip representation, we further develop a set aggregation cell to fuse the per frame features (see Fig.~\ref{fig:SA} for a block diagram). The set aggregation cell highlights the frame feature, with the aid of the clip feature, firstly, and then aggregates them by average pooling.

Let $[\Vec{f}^1, \ldots, \Vec{f}^t], \Vec{f}^j \in \mathbb{R}^{c}$ be a set of frame feature vectors, encoded by a deep network (see Fig.~\ref{fig:Strucutre}). The set aggregation cell first re-weights the frame features. In our implementation, we combine average pooling and max pooling to aggregate frame features. This is due to the fact that both pooling schemes encode different  statistical information and their combination is expected to increase the representation capacity. More specifically, each element in $\Vec{f}^{\mathrm{avg}}$ and $\Vec{f}^{\mathrm{max}}$ are defined as ${f}^{\mathrm{avg}}_{i} = \mathrm{avg}({f}^1_i, \ldots, {f}^t_i) = \frac{1}{t}\sum_{j = 1}^t({f}^j_i)$ and ${f}^{\mathrm{max}}_i = \mathrm{max}({f}^1_i, \ldots, {f}^t_i)$, respectively. Each aggregation is followed by self-gating layers (\ie, $\varpi$ and $\psi$ in Fig.~\ref{fig:SA}) to generate per-element modulation weights, and fused by  element-wise summation as: 

\begin{equation}
    \hat{\Vec{f}} = \varpi(\Vec{f}^\mathrm{avg}) \oplus \psi(\Vec{f}^\mathrm{max}).
\end{equation}

This is then followed by normalizing the fused weights to produce the final mask (\eg, $\Vec{m}_{\mathrm{s}} = \sigma(\hat{\Vec{f}})$) which is applied as follows:

\begin{equation}
\Mat{g}^j = \Mat{m}_{\mathrm{s}} \odot \Mat{f}^j,~j = 1, \ldots, t.
\end{equation}
Finally, we use average pooling to obtain the clip feature, $\Vec{g} = 1/t\sum_{j = 1}^t \Vec{g}^j$. We note that in our network the parameters in the two self-gating layers are not shared. This is to increase the diversity of features which is beneficial, and we evaluate it in \textsection~\ref{sec:expt}.

\begin{remark}
The set aggregation cell is inspired by the Squeeze-and-Excitation (SE) block~\cite{Hu_2018_CVPR}, in the sense that frame features will be emphasized under the context of the global clip-level features, but with a number of simple yet important differences: \textbf{(i)} The SE receives a feature map as input, while the input of our set aggregation is a set of frame features. \textbf{(ii)} The SE only uses global average pooling to encode the global feature of the feature maps, while the set aggregation employs both average and max pooling to encode hybrid clip features, exploiting more diverse information present in the frame features.
\end{remark}

\subsection{Implementation Details}
\textbf{Network Architecture and Training.}
We implemented our approach in the PyTorch~\cite{Pytorch} deep learning framework. We chose ResNet-50~\cite{He_2016_CVPR} as the backbone network, pre-trained on ImageNet \cite{russakovsky2015imagenet}. In a video clip with $t$ frames, each frame-level feature map, produced by the last convolutional layer, is squeezed to a feature vector $\Vec{f}^{j} \in \mathbb{R}^{2048}, j = 1, \ldots, t$ by global average pooling (GAP). Subsequently, the set aggregation cell fuses the frame features to a compact clip feature vector $\Vec{g}$. Following $\Vec{g}$, the final clip-level person representation $\Vec{F}$ is embedded by a fully connected (FC) layer with the dimension $1024$. Thereafter, another FC layer is added for the purpose of final classification during training. \textcolor{black}{In the channel recurrent attention module, the ratio $d$ is set to $16$ for the PRID-2011 and iLIDS-VID datasets, and $8$ for the MARS and DukeMTMC-VideoReID datasets, and the LSTM unit has one hidden layer. In the set aggregation cell, the self-gating layer is a bottleneck network to reduce the number of parameters, the dimension of the hidden vector is $2048/r$, and we choose $r = 16$ as in~\cite{Hu_2018_CVPR}, across all datasets. The ReLU and batch normalization are applied to each embedding layer and self-gating layer. The details of the datasets is described in \textsection\ref{dataset}.} 

We use the Adam~\cite{kingma2014adam} optimizer with default momentum. The initial learning rate is set to $3 \times 10^{-4}$ for PRID-2011 and iLIDS0-VID, and $4 \times 10^{-4}$ for MARS and DukeMTMC-VideoReID. The mini-batch size is set to $16$ for the PRID-2011 and iLIDS-VID datasets and $32$ for the MARS and DukeMTMC-VideoReID datasets, respectively. In a mini-batch, both $P$ and $K$ are set to $4$ for the PRID-2011 and iLIDS-VID, whereas $P = 8$, $K = 4$ for the MARS and DukeMTMC-VideoReID. The margin in the triplet loss, \ie, $\xi$, is set to $0.3$ for all datasets. The spatial size of the input frame is fixed to $256 \times 128$.  Following~\cite{gao2018revisiting}, $t$ is chosen as $4$ in all experiments and $4$ frames are \textit{randomly} sampled in each video clip~\cite{Zhao_2019_CVPR,gao2018revisiting}. Our training images are randomly flipped in the horizontal direction, followed by random erasing (RE)~\cite{zhong2017random}. We train the network for $800$ epochs. The learning rate decay is set to $0.1$, applied at the $200$-th, $400$-th epoch for the PRID-2011 and iLIDS-VID , and the $100$-th, $200$-th, $500$-th epoch for the MARS and DukeMTMC-VideoReID, respectively. Moreover, it is worth noting that we do not apply re-ranking to boost the ranking result in the testing phase.

\section{Experiment on Video Pedestrian Retrieval}\label{sec:expt}

\subsection{Datasets and Evaluation Protocol} \label{dataset}
In this section, we perform experiments on four standard video benchmark datasets, \ie, {PRID-2011}~\cite{PRID}, {iLIDS-VID}~\cite{WangTaiqing2016TPAMIiiLIDS-VID}, {MARS}~\cite{ZhengLiang2014ECCVMAR} and {DukeMTMC-VideoReID}~\cite{WuYu2018CVPROneShotforVideoReID} to verify the effectiveness of the proposed attentional network. \textbf{PRID-2011} has $400$ video sequences, showing $200$ different people where each person has $2$ video sequences, captured by two separate cameras. The person bounding box is manually labeled. \textbf{iLIDS-VID} contains $600$ image sequences of $300$ pedestrians, captured by two non-overlapping cameras in an airport. Each of the training and test sets has $150$ person identities. In this dataset, the target person is heavily occluded by other pedestrians or objects (\eg, baggage). \textbf{MARS} is one of the largest video person re-ID datasets which contains $1,261$ different identities and $20,715$ video sequences captured by $6$ separate cameras. The video sequences are generated by the GMMCP tracker~\cite{GMMCP}, and for each frame, the bounding box is detected by DPM~\cite{DPM}. The dataset is split into training and testing sets that contain $631$ and $630$ person identities, respectively. \textbf{DukeMTMC-VideoReID} is another large video person re-ID dataset. This dataset contains $702$ pedestrians for training, $702$ pedestrians for testing as well as $408$ pedestrians as distractors. The training set and testing set has $2,196$ video sequences and $2,636$ video sequences, respectively. The person bounding boxes are annotated manually.

Following existing works, we use both the cumulative matching characteristic (CMC) curve and mean average precision (mAP) to evaluate the performance of the trained re-ID system.

\subsection{Ablation Study}\label{ablationstudy}
This section demonstrates the effectiveness of the proposed blocks and the selection of appropriate hyper parameters via a thorough battery of experiments.

\noindent\textbf{Effect of Channel Recurrent Attention.} Here, we evaluate the effectiveness of the proposed channel recurrent attention, and verify our claim that our channel recurrent attention is able to capture more structure information as we sequentially feed the spatial vector to the LSTM. To show the design is reasonable, we compare our channel recurrent attention with two variations, namely, the spatial recurrent attention and the conv attention.

\begin{figure}[ht]
\centering
\subfigure[Channel recurrent attention.]{\includegraphics[width=6cm]{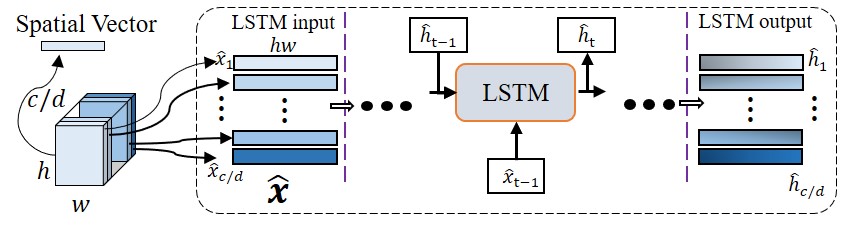}\label{cra_1}}%
\hfil
\subfigure[Spatial recurrent attention.]{\includegraphics[width=6cm]{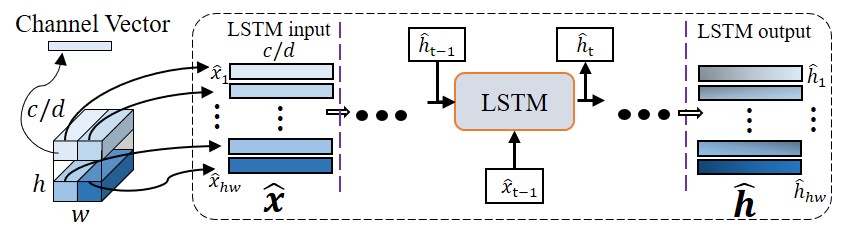}\label{sra_2}}%
\caption{\footnotesize{Schematic comparison between channel recurrent attention and spatial recurrent attention.}}\label{fig:diff_ra}
\end{figure}

In the spatial recurrent attention, the LSTM receives a sequence of channel features from feature maps as input, with the recurrent operator along the spatial domain. In more detail, in channel recurrent attention (see Fig.~\ref{fig:RA}), the input is a sequence of spatial vectors, (\eg, $\hat{\Mat{x}} = [\hat{{x}}_1, \ldots, \hat{{x}}_{\frac{{c}}{d}}]^{\top} \in \mathbb{R}^{\frac{{c}}{d} \times hw}$). In the spatial recurrent attention, the input is a sequence of channel vectors, (\eg, $\hat{\Mat{x}} = [\hat{{x}}_1, \ldots, \hat{{x}}_{hw}]^{\top} \in \mathbb{R}^{ hw \times \frac{{c}}{d}}$). Though the recurrent operation along the spatial domain is also able to learn the pattern spatially, the spatial recurrent attention lacks explicit modeling in the spatial domain. Fig.~\ref{fig:diff_ra} shows the schematic difference between channel recurrent attention (see Fig.~\ref{cra_1}) and spatial recurrent attention (see Fig.~\ref{sra_2}).  

\begin{figure}[ht]
\centering
\includegraphics[width = 0.7\textwidth]{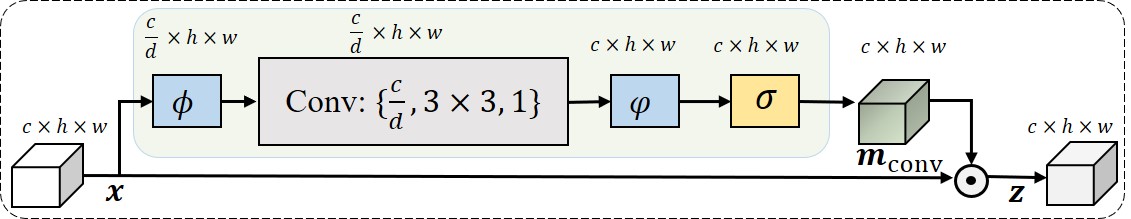}
\caption{\footnotesize{The architecture of the proposed conv attention module.}}
\label{fig:CAC}
\end{figure}

In addition, to verify the necessity of a global receptive field in our channel recurrent attention, we further replace the LSTM with a convolutional layer with a similar parameter size, which is called a conv attention. The architecture of the conv attention is shown in Fig.~\ref{fig:CAC}. In the $\mathrm{Conv}$ block, the kernel size is $3 \times 3$ and the sliding step is $1$, and it produces a tensor with the shape of $\frac{{c}}{d} \times h \times w$. The generated attention mask can be formulated as $\Mat{m}_{\mathrm{conv}} = {\sigma}\Big(\varphi\big(\mathrm{Conv}(\phi({\Mat{x}}))\big)\Big)$, where $\mathrm{Conv}(\cdot)$ indicates the convolutional operation.


\begin{table}[!ht]
\caption{\footnotesize{Comparison of three attention variations across four datasets. CRA: Channel Recurrent Attention; SRA: Spatial Recurrent Attention; CA: Conv Attention.}}
\begin{center}
\scalebox{0.66}{
\begin{tabular}{c|c|c|c|c|c|c|c|c|c}
\toprule[2pt]
\hline
\multicolumn{2}{c|}{}   & \multicolumn{2}{c|}{~~PRID-2011~~} & \multicolumn{2}{c|}{~~iLIDS-VID~~} & \multicolumn{2}{c|}{~~MARS~~}&
\multicolumn{2}{c}{DukeMTMC-VideoReID}\\
\hline
\multicolumn{2}{c|}{~~Method~~}&~~R-1~~&~~mAP~~&~~R-1~~&~~mAP~~&~~R-1~~&~~mAP~~&~~R-1~~&~~mAP~~\\
\hline
(\romannumeral1) & ~~No Attention~~ &85.4&91.0&80.0 &87.1 & 82.3 & 76.2&87.5&86.2\\
\hline
(\romannumeral2) & + CRA&\textbf{92.1}&\textbf{94.6}&\textbf{87.0}&\textbf{90.6}&\textbf{86.8}&\textbf{81.6}&\textbf{94.7}&\textbf{94.1}\\
(\romannumeral3) & + SRA&87.9&92.1&83.3&87.4&84.6&78.4&89.4&87.8\\
(\romannumeral4) & + CA&89.6&92.8&84.2&88.2&85.2&79.7&91.2&90.1\\
\hline
\hline
\toprule[2pt]
\end{tabular}
}
\end{center}\label{table:recurrent}
\end{table}

Table~\ref{table:recurrent} compares the effectiveness of three attention variations. It is shown that our channel recurrent attention has a superior performance over the other two variations. As can be observed, the channel recurrent attention cell improves the accuracy significantly across all four datasets. \textcolor{black}{This observation supports our assumption that the attention receives a performance gain from explicit modeling of the global receptive field in each slice of the feature maps.}

\noindent\textbf{Effect of the Position of Channel Recurrent Attention.} The position of the channel recurrent attention block affects the information in the spatial or the channel dimensions. We want to explore the rich spatial and channel information; thus, we only consider the feature maps from the middle of the deep network as input to channel recurrent attention (\ie, $\text{P}_1$, $\text{P}_2$, and $\text{P}_3$ in Fig.~\ref{fig:Strucutre}). The comparison is illustrated in Table~\ref{table:position}. It shows that the system receives a better gain when adding the channel recurrent attention module at position $\text{P}_2$, which aligns with our motivation that more spatial information is utilized in the feature maps. The works~\cite{WangXiaolong2017CVPRNonLocalNN,Pengfei_2019_ICCV} also present a similar observation. \textcolor{black}{When applying the attention in $\text{P}_1$, $\text{P}_2$ and $\text{P}_3$, the network performs at its best.}

\begin{table}[!ht]
\caption{\footnotesize{Effect of the position of channel recurrent attention across four datasets. CRA: Channel Recurrent Attention.}}
\begin{center}
\scalebox{0.66}{
\begin{tabular}{c|c|c|c|c|c|c|c|c|c}
\toprule[2pt]
\hline
\multicolumn{2}{c|}{}  & \multicolumn{2}{c|}{~~PRID-2011~~}& \multicolumn{2}{c|}{~~iLIDS-VID~~} & \multicolumn{2}{c|}{~~MARS~~}&
\multicolumn{2}{c}{DukeMTMC-VideoReID}\\
\hline
\multicolumn{2}{c|}{~~Position~~}&~~R-1~~&~~mAP~~&~~R-1~~&~~mAP~~&~~R-1~~&~~mAP~~&~~R-1~~&~~mAP~~\\
\hline
(\romannumeral1) &~~No Attention~~ &85.4&91.0&80.0&87.1&82.3&76.2&87.5&86.2\\
\hline
(\romannumeral2) & + CRA in $\text{P}_1$   &89.6&92.2&85.3&88.2&85.0&80.6&92.7&92.2\\
(\romannumeral3) & + CRA in $\text{P}_2$   &{91.0}&{94.4}&{86.7}&{90.2
}&{86.4}&{81.2}&94.2&93.4\\
(\romannumeral4) &+ CRA in $\text{P}_3$   &90.3&92.6&86.0&88.4&86.1&80.8&93.5&92.7\\
(\romannumeral5) &+ CRA in $\text{P}_1$\&$\text{P}_2$\&$\text{P}_3$&   \textbf{92.1}&\textbf{94.6}&\textbf{87.0}&\textbf{90.6}&\textbf{86.8}&\textbf{81.6}&\textbf{94.7}&\textbf{94.1}\\
\hline
\hline
\toprule[2pt]
\end{tabular}
}
\end{center}\label{table:position}
\end{table}

\noindent\textbf{Effect of Reduction Ratio $1/d$ in Channel Recurrent Attention.} The ratio $1/d$ in the embedding function $\phi(\cdot)$ (see Fig. \ref{fig:RA}) is to reduce the channel dimensionality of the input feature maps, consequently, reducing the sequence length input to the LSTM; thus, it is an important hyper-parameter in the channel recurrent attention. Table~\ref{table:reduction} reveals that the best performance is obtained when $d = 16$ for small-scale datasets and $d = 8$ for large-scale datasets. This could be due to the fact that training a network with a large amount of training samples is less prone to overfitting. Furthermore, this table also shows the fact that the LSTM has difficulties in modeling very long sequences (\eg smaller $d$ in Table~\ref{table:reduction}). However, when the sequences are too short (\eg, $d = 32$), the channel features are compressed, such that some pattern information is lost.    

\begin{table}[!ht]
\caption{\footnotesize{Effect of reduction ratio $1/d$ in channel recurrent attention across four datasets.}}
\begin{center}
\scalebox{0.68}{
\begin{tabular}{c|c|c|c|c|c|c|c|c|c}
\toprule[2pt]
\hline
\multicolumn{2}{c|}{}  & \multicolumn{2}{c|}{~~PRID-2011~~} & \multicolumn{2}{c|}{~~iLIDS-VID~~} & \multicolumn{2}{c|}{~~MARS~~} &
\multicolumn{2}{c}{DukeMTMC-VideoReID}\\
\hline
\multicolumn{2}{c|}{~~Reduction Ratio~~}&~~R-1~~&~~mAP~~&~~R-1~~&~~mAP~~&~~R-1~~&~~mAP~~&~~R-1~~&~~mAP~~\\
\hline
(\romannumeral1) &~~No Attention~~   &85.4&91.0&80.0&87.1&82.3&76.2&87.5&86.2\\
\hline
(\romannumeral2) &$d = 2$   &88.7&92.1&84.0&88.7&84.8&80.2&93.4&92.8\\
(\romannumeral3) &$d = 4$   &89.8&92.6&85.6&89.1&85.2&80.3&93.9&93.4\\
(\romannumeral4) &$d = 8$   &91.0&93.2&86.3&89.4&\textbf{86.8}&\textbf{81.6}&\textbf{94.7}&94.1\\
(\romannumeral5) &$d = 16$   &\textbf{92.1}&\textbf{94.6}&\textbf{87.0}&\textbf{90.6
}&85.5&80.7&94.3&\textbf{94.3}\\
(\romannumeral6) &$d = 32$   &91.0&93.8&82.7&88.9&84.3&79.8&93.2&93.4\\
\hline
\hline
\toprule[2pt]
\end{tabular}
}
\end{center}\label{table:reduction}
\end{table}

\noindent\textbf{Why using LSTM in the Channel Recurrent Attention?} \textcolor{black}{In our channel recurrent attention, we use the LSTM to perform the recurrent operation for the spatial vector. We observed that once the order of the spatial vectors is fixed, the recurrent operation in the LSTM is able to learn useful information along the channel dimension. We further investigated using Bi-LSTM to replace the LSTM in the attention and evaluate its performance. Compared with LSTM, the Bi-LSTM only brings a marginal/no performance gain across different datasets, whereas it almost doubles the number of parameters and FLOPs in the attention model. Please refer to \textsection 1 of the supplementary material for details of those experiments. These empirical experimental results support the use of a regular LSTM in our attention module.
}

\noindent\textbf{Effect of Set Aggregation.} Table~\ref{table:s-t-a} shows the effectiveness of set aggregation and the effectiveness of different pooling schemes in the set aggregation block. It is clear that the individual set aggregation improves the network performance and the combination of attention modules continues to increase the performance gain; showing that two attention modules mine complementary information in the network. Furthermore, all pooling schemes improve the results of the network, showing that the network receives gains from set aggregation. The combination of the average pooling and the max pooling scheme with non-sharing weights further shows its superiority over the individual average or max pooling schemes. This observation can be interpreted as the average and max pooled features have complementary information when encoding clip-level representations.

\begin{table}[!ht]
\caption{\footnotesize{Effect of set aggregation across four datasets. CRA: Channel Recurrent Attention, SA: Set Aggregation, $\dagger$: Sharing weights, $\ddagger$: Non-sharing weights. }}
\begin{center}
\scalebox{0.66}{
\begin{tabular}{c|c|c|c|c|c|c|c|c|c}
\toprule[2pt]
\hline
\multicolumn{2}{c|}{}  & \multicolumn{2}{c|}{~~PRID-2011~~}& \multicolumn{2}{c|}{~~iLIDS-VID~~} & \multicolumn{2}{c|}{~~MARS~~} &
\multicolumn{2}{c}{DukeMTMC-VideoReID}\\
\hline
\multicolumn{2}{c|}{~~Method~~}&~~R-1~~&~~mAP~~&~~R-1~~&~~mAP~~&~~R-1~~&~~mAP~~&~~R-1~~&~~mAP~~\\
\hline
(\romannumeral1) & ~~No Attention~~ &85.4&91.0&80.0 &87.1 & 82.3 & 76.2&87.5&86.2\\
\hline
(\romannumeral2) & + CRA&{92.1}&{94.6}&{87.0}&{90.6}&{86.8}&{81.6}&94.7&94.1\\
(\romannumeral3) & + SA (Average \& Max Pooling)&87.6&92.3&84.7&89.1&85.2&80.5&91.2&88.9\\
\hline
(\romannumeral4) & + CRA \& SA (Avg Pooling) &94.4&95.2&87.9&91.2&87.2&82.2&95.6&95.0\\
(\romannumeral5) & + CRA \& SA (Max Pooling)  &93.3&94.8&87.3&90.8&86.9&81.2&95.2&94.6\\
\hline
(\romannumeral6) & + CRA \& SA$^{\dagger}$ (Avg \& Max Pooling) &95.5&96.1&88.2&92.4&87.7&82.6&95.9&95.3\\
(\romannumeral7) & + CRA \& SA$^{\ddagger}$ (Avg \& Max Pooling) &\textbf{96.6}&\textbf{96.9}&\textbf{88.7}&\textbf{93.0
}&\textbf{87.9}&\textbf{83.0}&\textbf{96.3}&\textbf{95.5}\\
\hline
\hline
\toprule[2pt]
\end{tabular}
}
\end{center}\label{table:s-t-a}
\end{table}

\subsection{Comparison to the State-of-the-Art Methods} 

To evaluate the superiority of our deep attentional network, we continue to compare our results with the current state-of-the-art approaches, shown in Table~\ref{table:SOTA-on-video-reid} and Table~\ref{table:SOTA-on-video-reid_duke}.

\begin{table*}[!ht]
\caption{\footnotesize{Comparison with the SOTA methods on PRID-2011, iLIDS-VID and MARS datasets.}}\label{table:SOTA-on-video-reid}
\begin{center}
\scalebox{0.70}
{
\begin{tabular}{l|c|ccccc|ccccc|ccccc}
\toprule[2pt]
\hline
\multirow{2}{*}{{Method}}
&\multirow{2}{*}{Publication} 
&\multicolumn{5}{c|}{\begin{tabular}[c]{@{}c@{}}PRID-2011\end{tabular}}
&\multicolumn{5}{c|}{\begin{tabular}[c]{@{}c@{}}iLIDS-VID\end{tabular}} 
&\multicolumn{5}{c}{\begin{tabular}[c]{@{}c@{}}MARS\end{tabular}}\\
\cline{3-17}
&&R-1	&R-5 &R-10 &R-20&mAP &R-1 &R-5 &R-10&R-20 &mAP&R-1 &R-5 &R-10&R-20&mAP \\
\hline
RFA-Net~\cite{RFA-Net}&ECCV'16&58.2&85.8&93.4&97.9&-&49.3&76.8&85.3&90.0&-&-&-&-&-&-\\
\hline
McLaughlin~\etal~\cite{McLaughlin2016CVPRRNNforVideoReID}&CVPR'16&70.0&90.0&95.0&97.0&-&58.0&84.0&91.0&96.0&-&-&-&-&-&-\\
\hline
ASTPN~\cite{XuShuangjie2017ICCVASTPN}&ICCV'17&77.0&95.0&99.0&99.0&-&62.0&86.0&94.0&98.0&-&44.0&70.0&74.0&81.0&-\\
\hline
MSCAN~\cite{Li_2017_CVPR}&CVPR'17&-&-&-&-&-&-&-&-&-&-&71.8&86.6&-&93.1&56.1\\
\hline
CNN+XQDA~\cite{ZhengLiang2014ECCVMAR}&ECCV'16&77.3&93.5&-&{99.3}&-&53.1&81.4&-&{95.1}&-&65.3&82.0&-&89.0&47.6\\
\hline
Zhou~\etal~\cite{Zhou2017SeeTF}&CVPR'17&79.4&94.4&-&99.3&-&55.2&86.5&-&97.0&-&70.6&90.0&-&97.6&50.7\\ 
\hline
Chen~\etal~\cite{ChenDapeng2018CVPRVideoPersonReID}&CVPR'18&88.6&99.1&-&-&90.9&79.8&91.8&-&-&82.6&81.2&92.1&-&-&69.4\\
+ Optcal flow& &93.0&99.3&100.0&100.0&94.5&85.4&96.7&98.8&99.5&{87.8}&86.3&94.7&-&98.2&76.1
\\
\hline
QAN~\cite{liu_2017_qan}&CVPR'17&90.3&98.2&99.3&100.0&-&68.0&86.8&-&97.4&-&73.7&84.9&-&91.6&51.7\\
\hline
Li~\etal~\cite{LiShuang2018CVPRSpatiotemporalAttentionforVideoReID}&CVPR'18&93.2&-&-&-&-&80.2&-&-&-&-&82.3&-&-&-&65.8\\
\hline
Gao~\etal~\cite{gao2018revisiting}&BMVC'18&-&-&-&-&-&-&-&-&-&-&83.3&93.8&96.0&97.4&76.7\\
\hline
PBR~\cite{Suh_2018_ECCV}&ECCV'18&-&-&-&-&-&-&-&-&-&-&83.0&92.8&95.0&96.8&72.2\\
\hline
SCAN~\cite{ZhangRuimiao2018arXivSCANforVideoReID}&TIP'19&92.0&98.0&100.0&100.0&-&81.3&93.3&96.0&98.0&-&86.6&94.8&-&{98.1}&{76.7}\\
+ Optical flow & &95.3&99.0&100.0&100.0&-&{88.0}&96.7&98.0&{100.0}&-&{87.2}&{95.2}&-&{98.1}&77.2\\
\hline
STIM-RRU~\cite{Liu2019SpatialAT}&AAAI'19&92.7&98.8&-&99.8&-&84.3&96.8&-&{100.0}&-&84.4&93.2&-&96.3&72.7\\
\hline
COSAM~\cite{COSAM_2019_ICCV}&ICCV'19&-&-&-&-&-&79.6 & 95.3 & - &-&- & 84.9 &95.5 &-&97.9 & 79.9\\
\hline
STAR+Optical flow ~\cite{BMVC2019STAR}&BMVC'19&93.4&98.3&100.0&100.0&-&85.9&{97.1}&{98.9}&{99.7}&-&85.4&{95.4}&96.2&97.3&76.0\\
\hline
STA~\cite{Fu2018STASA}& AAAI'19 &-&-&-&-&-&-&-&-&-&-&86.3&95.7&-&98.1&80.8\\
\hline
VRSTC~\cite{Hou_2019_CVPR_VRSTC}&CVPR'19&-&-&-&-&-&83.4&95.5&97.7&99.5&-&\textbf{88.5}&96.5&97.4&-&82.3\\
\hline
Zhao \etal~\cite{Zhao_2019_CVPR}&CVPR'19&93.9&99.5&-&100.0&-&{86.3}&{97.4}&-&{99.7}&-&{87.0}&95.4&-&98.7&78.2\\
\hline
GLTR~\cite{Li_2019_ICCV}&ICCV'19&95.5&\textbf{100.0}&-&-&-&86.0&{98.0}&-&-&-&87.0&95.7&-&98.2&78.4\\
\hline
Baseline&-&85.4&98.9&98.9&98.9&91.0&80.0&95.3&98.7&99.3&87.1&82.3&93.9&95.8&97.2&76.2\\
Ours&-&\textbf{96.6}&98.9&\textbf{100.0}&\textbf{100.0}&\textbf{96.9}&\textbf{88.7}&\textbf{97.3}&\textbf{99.3}&\textbf{100.0}&\textbf{93.0}&87.9&\textbf{96.6}&\textbf{97.5}&\textbf{98.8}&\textbf{83.0}\\
\hline
\hline
\toprule[2pt]
\end{tabular}
}
\end{center}
\end{table*}

\begin{table*}[ht]
\caption{\footnotesize{Comparison with the SOTA methods on DukeMTMC-VideoReID  dataset.}}\label{table:SOTA-on-video-reid_duke}
\begin{center}
\scalebox{0.68}
{
\begin{tabular}{l|c|ccccc}
\toprule[2pt]
\hline
\multirow{2}{*}{{~~Method}~~}
&\multirow{2}{*}{~~Publication~~} 
&\multicolumn{5}{c}{\begin{tabular}[c]{@{}c@{}}~~~~~DukeMTMC-VideoReID~~~~~\end{tabular}}\\
\cline{3-7}
&&~~R-1~~&~~R-5~~&~~R-10~~&~~R-20~~&~~mAP~~\\
\hline
ETAP-Net~\cite{WuYu2018CVPROneShotforVideoReID}&CVPR'18&83.6&94.6&-&97.6&78.3\\
STAR+Optical flow~\cite{BMVC2019STAR}&BMVC'19 &94.0&99.0&99.3&99.7&93.4\\
VRSTC~\cite{Hou_2019_CVPR_VRSTC}&CVPR'19&95.0&99.1&99.4&-&93.5\\
STA~\cite{Fu2018STASA}& AAAI'19&96.2&99.3&-&99.7&94.9\\
GLTR~\cite{Li_2019_ICCV}&ICCV'19&\textbf{96.3}&99.3&-&99.7&93.7\\
\hline
Baseline&-&87.5&96.5&97.2&98.3&86.2\\
Ours&-&\textbf{96.3}&\textbf{99.4}&\textbf{99.7}&\textbf{99.9}&\textbf{95.5}\\
\hline
\hline
\toprule[2pt]
\end{tabular}
}
\end{center}
\end{table*}

\noindent\textbf{PRID-2011.} On the PRID-2011 dataset, our network improves the state-of-the-art accuracy by $1.1\%$ in R-1, compared to GLTR~\cite{Li_2019_ICCV}. As for the mAP, our approach outperforms \cite{ChenDapeng2018CVPRVideoPersonReID} by $2.4\%$. When compared to SCAN~\cite{ZhangRuimiao2018arXivSCANforVideoReID}, which uses optical flow, our approach outperforms it by $1.3\%$ in R-1.  

\noindent\textbf{iLIDS-VID.} On the iLIDS-VID dataset, our approach improves the state-of-the-art mAP value by \textcolor{black}{$5.2\%$}, compared to~\cite{ChenDapeng2018CVPRVideoPersonReID}. As for the R-1 accuracy, our approach also achieves a new state-of-the-art, outperforming \cite{Zhao_2019_CVPR} by a comfortable \textcolor{black}{$2.4\%$}. In addition, our approach continues to outperform SCAN + optical flow~\cite{ZhangRuimiao2018arXivSCANforVideoReID} by $0.7\%$ in R-1.

\noindent\textbf{MARS.} On the MARS dataset, our approach achieves state-of-the-art performances on mAP and competitive performance on the CMC curve. In particular, our approach outperforms VRSTC~\cite{Hou_2019_CVPR_VRSTC} on mAP, R-5 and R-10. It is worth mentioning that VRSTC uses a generator for data augmentation. Furthermore, when compared to other methods, we observe that our approach outperforms GLTR~\cite{Li_2019_ICCV} by $1.3\%$/$4.6\%$ in R-1/mAP.

\noindent\textbf{DukeMTMC-VideoReID.} As for this new dataset, our network continues to show its superior performance (see Table~\ref{table:SOTA-on-video-reid_duke}). Our approach is superior to GLTR by $1.8\%$ on mAP, and outperform the state-of-the-art mAP value of STA by $0.6\%$, and our network also achieves competitive performance on the CMC metric, outperforming the state-of the-art on R-5, R-10 and R-20.

\begin{figure}[ht] 
\centering
\includegraphics[width = 0.9\linewidth]{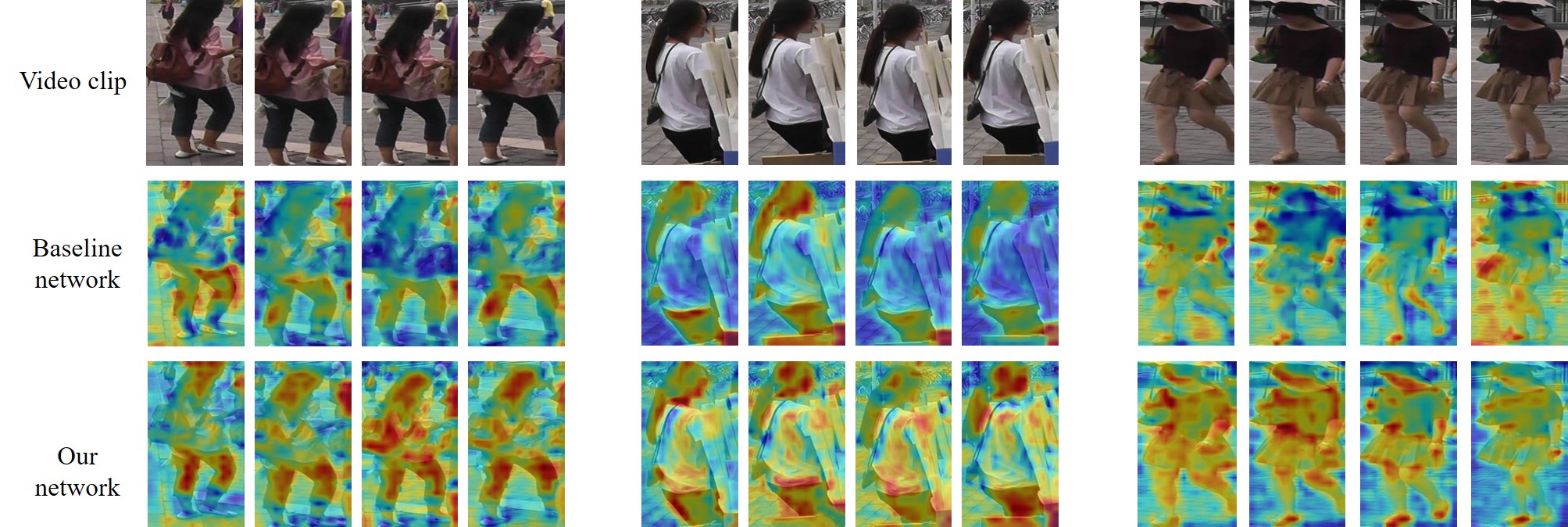}
\caption{\footnotesize{Visualization of feature maps. We sample three video clips from different pedestrians and visualize the feature maps.}} \label{fig:Vis}
\end{figure}

We visualize the feature maps from the baseline network and our channel recurrent attention network, trained on the MARS dataset. The feature maps are obtained in $\text{P}_2$ (see Fig.~\ref{fig:Strucutre}).  In Fig.~\ref{fig:Vis}, we observed that compared to the baseline network, our attention network highlights more areas of human bodies, which verifies the effectiveness of our network qualitatively. Please refer to the supplementary material for further visualizations.      

\section{Experiments on Single Image Pedestrian Retrieval}\label{sec:exten}
To show the generalisation of the proposed channel recurrent attention, we employ it in a single image pedestrian retrieval task. We select a strong baseline network from~\cite{Pengfei_2019_ICCV}, and insert the channel recurrent attention after each convolutional block. The deep network is fine-tuned from ImageNet pre-training~\cite{russakovsky2015imagenet} and trained with the same hyper-parameter setting as in~\cite{Pengfei_2019_ICCV}. We use \textbf{CUHK01}~\cite{LiWei2012ACCVCUHK01} and \textbf{DukeMTMC-reID}~\cite{ristani2016MTMC} to evaluate the performance of the network. Please refer to the supplementary material for the details of the datasets.

We use mAP and the CMC curve to evaluate the performance. Table~\ref{table:cuhk01} and Table~\ref{table:dukeimage} illustrate that our approach achieves competitive results to existing state-of-the-art approaches, showing the effectiveness and generalization of our channel recurrent attention module.

\begin{table*}[!ht]
\parbox{.46\linewidth}{
\centering
\caption{\small{Comparison with the SOTA on CUHK01  dataset.}}\label{table:cuhk01}
\resizebox{0.46\textwidth}{!}{
\begin{tabular}{l|c|cccc}
\toprule[2pt]
\hline
\multirow{2}{*}{{~~Method}~~}
&\multirow{2}{*}{~~Publication~~} 
&\multicolumn{4}{c}{\begin{tabular}[c]{@{}c@{}}~~~~~CUHK01~~~~~\end{tabular}}\\
\cline{3-6}
&&~~R-1~~&~~R-5~~&~~R-10~~&~~R-20~~\\
\hline
Zhao~\etal~\cite{Zhao_2017_ICCV_Deeply} &ICCV'17&75.0&93.5&95.7&97.7\\
Spindle Net~\cite{Zhao_2017_CVPR}&CVPR'17&79.9&94.4&97.1&98.6\\
PBR~\cite{Suh_2018_ECCV}&ECCV‘18&80.7&94.4&97.3&98.6\\
\hline
Baseline    &-&79.3&92.7&95.8&98.2 \\
Ours        &-&\textbf{83.3}&\textbf{96.3}&\textbf{98.4}&\textbf{98.9}\\
\hline
\hline
\toprule[2pt]
\end{tabular}
}
}
\hfill
\parbox{.46\linewidth}{
\centering
\caption{\small{Comparison with the SOTA on DukeMTMC-reID dataset.}}\label{table:dukeimage}
\resizebox{0.46\textwidth}{!}{
\begin{tabular}{l|c|cccc}
\toprule[2pt]
\hline
\multirow{2}{*}{{~~Method}~~}
&\multirow{2}{*}{~~Publication~~} 
&\multicolumn{4}{c}{\begin{tabular}[c]{@{}c@{}}~~~~~DukeMTMC-reID~~~~~\end{tabular}}\\
\cline{3-6}
&&~~R-1~~&~~R-5~~&~~R-10~~&~~mAP~~\\
\hline
OS-Net~\cite{OSNET}&ICCV'19&88.6&-&-&73.5\\
BAT-net~\cite{Pengfei_2019_ICCV}&ICCV'19&87.7&94.7&96.3&77.3\\
ABD-Net~\cite{Chen_2019_ICCV_ABD} &ICCV'19&89.0&-&-&\textbf{78.6}\\
\hline
Baseline     &-&85.4&93.8&95.5&75.0\\
Ours         &-&\textbf{89.2}&\textbf{95.6}&\textbf{96.9}&78.3\\
\hline
\hline
\toprule[2pt]
\end{tabular}
}
}
\end{table*}

\section{Conclusion}\label{sec:conclusion}
This work proposes a novel deep attentional network for task of video pedestrian retrieval. This network benefits from the developed channel recurrent attention and set aggregation modules. The channel recurrent attention module is employed for a global view to feature maps, to learn the channel and spatial pattern jointly, given a frame feature maps as input. Then the set aggregation cell continues to re-weight each frame feature and fuses them to get a compact clip representation. Thorough evaluation shows that the proposed deep network achieves state-of-the-art results across four standard video-based person re-ID datasets, and the effectiveness of each attention is further evaluated by extensive ablation studies.  



\bibliographystyle{splncs}
\bibliography{egbib}

\clearpage
\appendix
\section{Channel Recurrent Attention Module Analysis}

In the channel recurrent attention module, we use an LSTM to jointly capture spatial and channel information. In the implementation, we feed the spatial vectors to the LSTM sequentially, such that the recurrent operation of the LSTM captures the channel pattern while the FC layer in the LSTM has a global receptive field of each spatial slice. Since the LSTM is a temporal model and its output depends on the order of the input sequences, we analyze how the order of the input spatial vectors affects the attention performance and whether the LSTM has the capacity to learn the pattern in the channel dimension of the feature maps.

In the main paper, we feed the spatial vectors to the LSTM sequentially in a \say{\textbf{forward}} manner (\ie, from $\hat{{x}}_1$ to $\hat{{x}}_{\frac{c}{d}}$). We continue to define other configurations to verify how the sequence order affects the attention performance. The \say{\textbf{reverse}} configuration: feeding the spatial vectors from $\hat{{x}}_{\frac{c}{d}}$ to $\hat{{x}}_1$ to the LSTM, whose direction is opposite to that in the \say{forward} configuration. In the \say{\textbf{random shuffle}} configuration, we first randomly shuffle the order of spatial vectors in $\hat{\Vec{x}}$, and then feed them to the LSTM sequentially. Then we recover the produced $\hat{\Vec{h}}$ and generate the attention maps. This \say{random shuffle} is operated in each iteration during training. The last configuration, we term \say{\textbf{fixed permutation}}. In this configuration, we randomly generate a permutation matrix (\ie, $\Vec{p}_1$) and apply to $\hat{\Vec{x}}$, to produce $\hat{\Vec{x}}^p$ (\ie, $\hat{\Vec{x}}^p = \Vec{p}_1\hat{\Vec{x}}$). Then we feed each row of $\hat{\Vec{x}}^p$ to the LSTM and obtain $\hat{\Vec{h}}^p$ and apply another permutation matrix $\Vec{p}_2$, as $\hat{\Vec{h}} = \Vec{p}_2 \hat{\Vec{h}}^p$. Here, $\Vec{p}_2 = \Vec{p}_1^{\top}$ and $\Vec{p}_1$, $\Vec{p}_2$ are fixed during training. For this configuration, we perform the experiments twice with two different permutation matrices.

We empirically compare the aforementioned four configurations on the iLIDS-VID and the MARS datasets, shown in Table~\ref{table:CRAM}. From Table~\ref{table:CRAM}, we observe that the LSTM does indeed learn useful information along the channel dimension via the recurrent operation (\ie, row (\romannumeral2), (\romannumeral4), (\romannumeral5) and (\romannumeral6)) when the order of the spatial vectors is fixed during training. However, if we randomly shuffle the order of the spatial vectors before feeding to them to the LSTM in each iteration (\ie, row (\romannumeral3)), the LSTM fails to capture useful information in the channel, and the attention mechanism even degrades below the performance of the baseline network on the MARS dataset (\ie, row (\romannumeral1)).

In this analysis, we can draw the conclusion that the order of the spatial vectors has a minor influence on the attention performance when the order of spatial vectors is fixed. However, it is still difficult to figure out the optimal order of spatial vectors. In all experiments, we empirically use the \say{forward} configuration in our attention mechanism.

\begin{table}[!ht]
\caption{\footnotesize{Channel recurrent attention module analysis on the iLIDS-VID~\cite{WangTaiqing2016TPAMIiiLIDS-VID} and the MARS~\cite{ZhengLiang2014ECCVMAR} datasets.}}
\begin{center}
\scalebox{1}{
\begin{tabular}{c|c|c|c|c|c}
\toprule[2pt]
\hline
\multicolumn{2}{c|}{}  & \multicolumn{2}{c|}{~~iLIDS-VID~~} & \multicolumn{2}{c}{~~MARS~~} \\
\hline
\multicolumn{2}{c|}{~~Sequences order~~}&~~R-1~~&~~mAP~~&~~R-1~~&~~mAP~~\\
\hline
(\romannumeral1) & ~~No Attention~~&80.0 &87.1 & 82.3 & 76.2\\
\hline
(\romannumeral2) & ~~Forward~~&87.0&90.6&86.8&81.6\\
(\romannumeral3) & ~~Reverse~~&86.8&90.7&86.3&81.2\\
(\romannumeral4) & ~~Random Shuffle~~&82.7&88.8&79.2&72.4\\
(\romannumeral5) & ~~Fixed Permutation~$\mathrm{1}$~~&86.4&89.3&85.8&80.4\\
(\romannumeral6) & ~~Fixed Permutation~$\mathrm{2}$~~&86.7&90.3&86.1&80.9\\
\hline
\hline
\toprule[2pt]
\end{tabular}
}
\end{center}\label{table:CRAM}
\end{table}

\textcolor{black}{Intuitively, we further use Bi-LSTM to replace the LSTM in the channel recurrent attention module, to verify whether the sophisticated recurrent network is able to learn more complex information in the channel dimension. Table~\ref{table:Comparison} compares the difference of LSTM and Bi-LSTM in channel recurrent attention module. This study shows that the attention w/ Bi-LSTM cannot brings more performance gain than the that w/ LSTM. However, the Bi-LSTM almost doubles the computation complexities and parameters. Thus we choose regular LSTM in our attention module. 
}

\begin{table}[!ht]
\caption{\footnotesize{Comparison of LSTM and Bi-LSTM in channel recurrent attention module on the iLIDS-VID~\cite{WangTaiqing2016TPAMIiiLIDS-VID} and the MARS~\cite{ZhengLiang2014ECCVMAR} datasets. FLOPs: the number of FLoating-point OPerations, PNs: Parameter Numbers.}}
\begin{center}
\scalebox{0.9}{
\begin{tabular}{c|c|c|c|c|c|c|c}
\toprule[2pt]
\hline
\multicolumn{2}{c|}{}  & \multicolumn{2}{c|}{~~iLIDS-VID~~} & \multicolumn{2}{c|}{~~MARS~~} & \multicolumn{2}{c}{~~Comparison~~}\\
\hline
\multicolumn{2}{c|}{~~Model~~}&~~R-1~~&~~mAP~~&~~R-1~~&~~mAP~~&~~FLOPs~~&~~PNs~~\\
\hline
(\romannumeral1) & ~~No Attention~~&80.0 &87.1 & 82.3 & 76.2&~~$3.8 \times 10^9$~~&~~$25.4 \times 10^6$~~\\
\hline
(\romannumeral2) & ~~CRA w/LSTM~~&87.0&90.6&86.8&81.6&$0.18 \times 10^9$&$2.14 \times 10^6$\\
(\romannumeral2) & ~~CRA w/ Bi-LSTM~~&87.2&90.2&85.4&81.0&$0.32 \times 10^9$&$4.25 \times 10^6$\\
\hline
\hline
\toprule[2pt]
\end{tabular}
}
\end{center}\label{table:Comparison}
\end{table}

\section{Set Aggregation Cell Analysis}
In this section, we show the analysis of modeling the video clip as a set and the set aggregation cell acting as a valid set function.

In our channel recurrent attention network, we sample $t$ frames in a video sequence \textit{randomly}, to construct a video clip (\ie, $[T^1, \ldots, T^t], T^j \in \mathbb{R}^{C\times H \times W}$) with its person identity as label (\ie, $y$). In such a video clip, the frames are order-less and the order of frames does not affect the identity prediction by the network during training. The video frames are fed to the deep network and encoded to a set of frame feature vectors (\ie, $\Vec{F} = [\Vec{f}^1, \ldots, \Vec{f}^t], \Vec{f}^j \in \mathbb{R}^{c}$), then the frame features are fused to a discriminative clip representation (\ie, $\Vec{g}$) by the aggregation layer (\ie, set aggregation cell).

The set aggregation cell realizes a permutation invariant mapping, $g_{\kappa}: \mathcal{F} \to \mathcal{G}$ from a set of vector spaces onto a vector space, such that the frame features (\ie, $\Vec{F} = [\Vec{f}^1, \ldots, \Vec{f}^t], \Vec{f}^j \in \mathbb{R}^{c}$) are fused to a compact clip representation (\eg, $\Vec{g} \in \mathbb{R}^{c}$). If in this permutation invariant function (\ie, $g_{\kappa}$), the input is a set, then the response of the function is invariant to the ordering of the elements of its input. This property is described as:

\begin{property}~\cite{NIPS2017_6931_deepset}
A function $g_{\kappa}: \mathcal{F} \to \mathcal{G}$ acting on sets must be \textbf{invariant} to the order of objects in the set, \ie, for any permutation $\Pi: g_{\kappa}\big( [\Vec{f}^1, \ldots, \Vec{f}^t] \big) = g_{\kappa}\big( [\Vec{f}^{\Pi(1)}, \ldots, \Vec{f}^{\Pi(t)}]\big)$.
\end{property}

\begin{figure}[ht] 
\centering
\includegraphics[width = 0.45\linewidth]{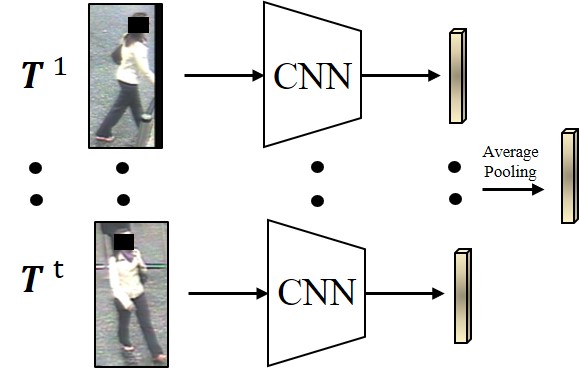}
\caption{\footnotesize{The pipeline of fusing frame features. The frame features are fused by using average pooling; thus the pedestrian identity predictor is permutation invariant to the order of frames in a clip.}} \label{fig:fff}
\end{figure}

In our supervised video pedestrian retrieval task, it is given $t$ frame samples of $T^1, \ldots, T^t$ as well as the person identity $y$. Since the frame features are fused using average pooling, shown in Fig.~\ref{fig:fff}, thus it is obvious that the pedestrian identity predictor is permutation invariant to the order of frames in a clip (\ie, $f_{\theta}([T^1, \ldots, T^t]) = f_{\theta}([T^{\Pi(1)}, \ldots, T^{\Pi(t)}])$ for any permutation $\Pi$). We continue to study the structure of the set function on \textit{countable sets} and show that our set aggregation cell satisfies the structure of the set function.

\begin{theorem}~\cite{NIPS2017_6931_deepset}
Assume the elements are countable, \ie, $|\mathfrak{X}| < \mathfrak{N}_0$. A function $g_{\kappa}: 2^{\mathfrak{X}} \to \mathbb{R}^c$, operating on a set $\Vec{F} = [\Vec{f}^1, \ldots, \Vec{f}^t]$ can be a valid set function, \ie it is permutation invariant to the elements in $\Vec{F}$, if and only if it can be decomposed in the form $\beta\big(\sum_{\Vec{f} \in \Vec{F}}\gamma(\Vec{f})\big)$, for suitable transformations $\beta$ and $\gamma$.
\end{theorem}

\begin{figure}[!ht]
\centering
\subfigure[Set aggregation cell with $\gamma$ containing the $\mathrm{Avg}$ function.]{\includegraphics[width=9.2cm,height=1.9cm]{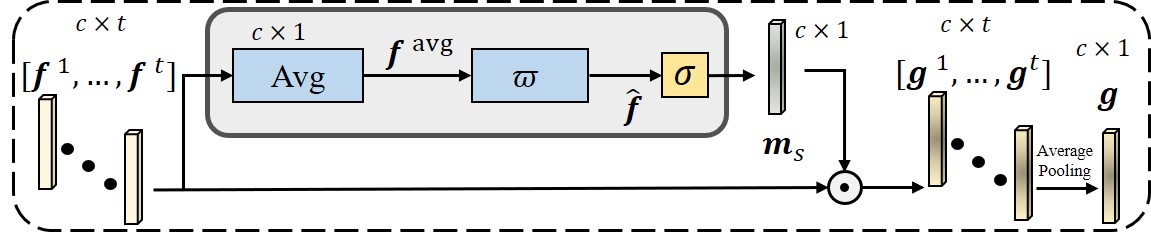}\label{formavg}}%
\hfil
\subfigure[Set aggregation cell with $\gamma$ containing the $\mathrm{Max}$  function.]{\includegraphics[width=9.2cm,height=1.9cm]{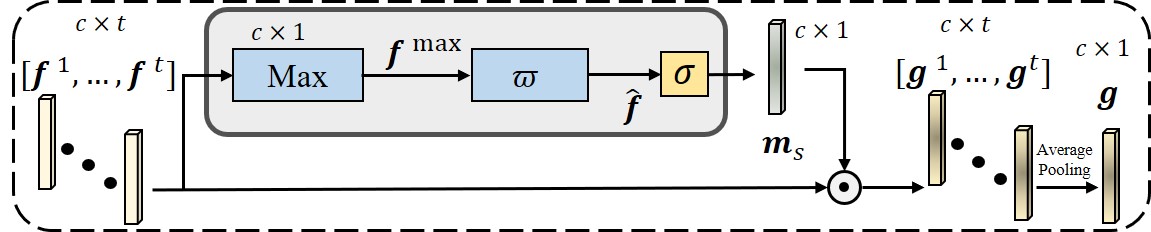}\label{formmax}}%
\caption{\footnotesize{Two set aggregation cells following from $\beta\big(\sum_{\Vec{f} \in \Vec{F}}\gamma(\Vec{f})\big)$.}}\label{fig:form1}
\end{figure}

In our deep architecture, we use an aggregation layer (\ie, set aggregation cell) to fuse frame features in a countable set (\ie, $|F| = t$), and this aggregation layer is a permutation invariant function. We use a simple case as an example, shown in Fig.~\ref{formavg}. In this architecture, the $\gamma$ function is a mapping: $\mathbb{R}^{c \times t} \to \mathbb{R}^{c \times t}$, formulated as:
\begin{equation}
\Vec{G} = \gamma (\Vec{F}) = \sigma\Big(\varpi\big(\mathrm{Avg}(\Vec{F})\big)\Big) \odot \Vec{F},
\end{equation}
where $\Vec{G} = [\Vec{g}^1, \ldots, \Vec{g}^t]$ and $\Vec{F} = [\Vec{f}^1, \ldots, \Vec{f}^t]$. Thereafter, average pooling operates on the feature set, to realize the summation and $\beta$ function. Since the $\gamma$ and $\beta$ functions are all permutation invariant, the set aggregation cell is a valid set function. Similarly, in Fig.~\ref{formmax}, the $\gamma$ function is realized as:
\begin{equation}
\Vec{G} = \gamma(\Vec{F}) = \sigma\Big(\varpi\big(\mathrm{Max}(\Vec{F})\big)\Big) \odot \Vec{F},
\end{equation}
which also satisfies the condition of permutation invariance of its input. In the main paper, we evaluated the performance of two vanilla aggregation cells empirically and we observed that the aggregation cell with $\mathrm{Avg}$ function is superior to that with the $\mathrm{Max}$ function.

\begin{figure}[!ht] 
\centering
\includegraphics[width=9.2cm,height=1.9cm]{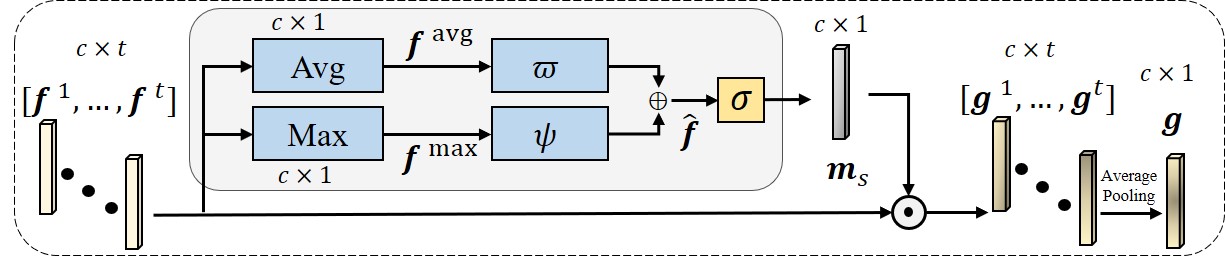}
\caption{\footnotesize{The architecture of the proposed set aggregation cell in the main paper.}} \label{fig:SA_s}
\end{figure}

Since the $\mathrm{Avg}$ and the $\mathrm{Max}$ operations are permutation invariant, their summation is also permutation invariant; thus we continue to develop our set aggregation in the main paper, shown in Fig.~\ref{fig:SA_s}. The $\gamma$ function is formulated as:
\begin{equation}
\Vec{G} = \gamma(\Vec{F}) = \sigma\Big(\varpi\big(\mathrm{Avg}(\Vec{F})\big) \oplus \psi \big(\mathrm{Max}(\Vec{F})\big) \Big) \odot \Vec{F}.
\end{equation}

The above analysis shows the necessity to model the frame features in a clip as a set and that the set aggregation cell is a valid set function. In the main paper, we also verify the effectiveness of the set aggregation cell.

\section{Loss Function}\label{sec:loss}

\noindent\textbf{Triplet Loss.} To take into account the between-class variance, we use the triplet loss~\cite{SchroffFlorian2015CVPRFaceNet}, denoted $\mathcal{L}_{\text{tri}}$, to encode the relative similarity information in a triplet. In a mini-batch, a triplet is formed as $\{\Mat{T}_i, \Mat{T}_i^{+}, \Mat{T}_i^{-}\}$, such that the anchor clip $\Mat{T}_i$ and the positive clip $\Mat{T}_i^{+}$ have the same identity, while the negative clip $\Mat{T}_i^{-}$ has a different identity. With the clip feature embedding, the triplet loss is formulated as: $\mathcal{L}_{\text{tri}} = \frac{1}{PK}\sum_{i = 1}^{PK}\Big[\|\Vec{F}_i - \Vec{F}_i^+\| - \|\Vec{F}_i - \Vec{F}^-_i\| + \xi \Big]_+$, where $\xi$ is a margin and $[\cdot]_+ = \mathrm{max}(\cdot, 0)$. A mini-batch is constructed by randomly sampling $P$ identities and $K$ video clips for each identity. We employ a hard mining strategy~\cite{HermansAlexander2017arXivInDefenseoftheTripletLoss} for triplet mining.   

\noindent\textbf{Cross-entropy Loss.} The cross-entropy loss realizes the classification task in training a deep network. It is expressed as: $\mathcal{L}_{\text{sof}} = \frac{1}{PK}\sum_{i = 1}^{PK}-\mathrm{log}\big(p(y_i|\Vec{F}_{i})\big)$, where $p$ is the predicted probability that $\Vec{F}_{i}$ belongs to identity $y_i$. The classification loss encodes the class specific information, which minimizes the within-class variance. The total loss function is formulated as: $\mathcal{L}_{\text{tot}} = \mathcal{L}_{\text{sof}} + \mathcal{L}_{\text{tri}}$.

\section{Description of Image Pedestrian Datasets}
In the main paper, we evaluate our attention module on image person re-ID tasks on the CUHK01~\cite{LiWei2012ACCVCUHK01} and the DukeMTMC-reID~\cite{ristani2016MTMC} datasets. The description of the two datasets is as follows:

\noindent\textbf{CHUK01} contains $3,884$ images of $971$ identities. The person images are collected by two cameras with each person having two images per camera view (\ie, four images per person in total). The person bounding boxes are labelled manually. We adopt the $485$/$486$ training protocol to evaluate our network. 

\noindent\textbf{DukeMTMC-reID} is the image version of DukeMTMC-VideoReID dataset for the re-ID purpose. It has $1,404$ identities and includes $16,522$ training images of $702$ identities, $2,228$ query and $17,661$ gallery images of $702$ identities. The pedestrian bounding boxes are labeled manually.

We use a single query (SQ) setting for both datasets when calculating the network prediction accuracy.

\section{Pedestrian Samples of Datasets}
\begin{figure}[!ht]
\centering
\subfigure[Samples from the PRID-2011 dataset.]{
\includegraphics[width=5.4cm]{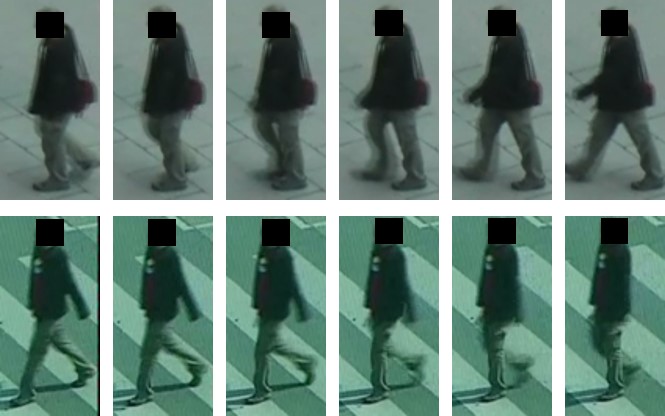}
\label{samp_prid}
}
\quad
\subfigure[Samples from the iLIDS-VID dataset.]{
\includegraphics[width=5.4cm]{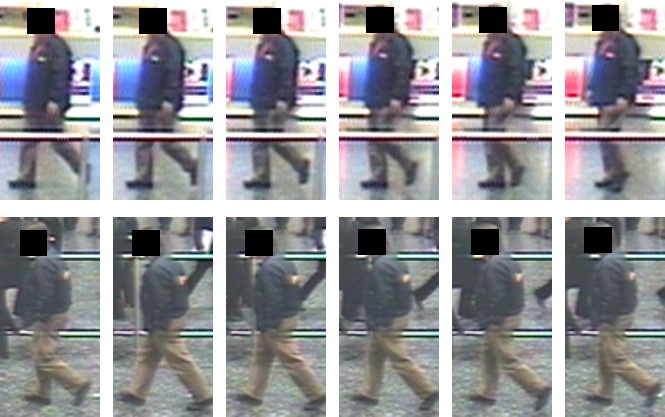}
\label{samp_ilid}
}
\quad
\subfigure[Samples from the MARS dataset.]{
\includegraphics[width=5.4cm]{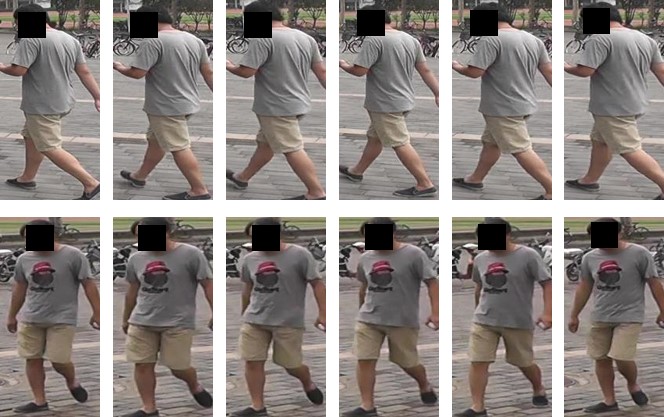}
\label{samp_mars}
}
\quad
\subfigure[Samples from the DukeMTMC-VideoReID]{
\includegraphics[width=5.4cm]{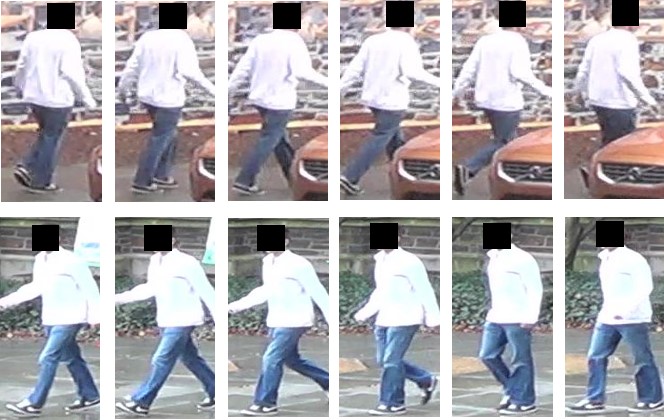}
\label{samp_duke}
}
\caption{\small{Samples from: (a) PRID-2011 dataset~\cite{PRID}, (b) iLIDS-VID dataset~\cite{WangTaiqing2016TPAMIiiLIDS-VID}, (c) MARS dataset~\cite{ZhengLiang2014ECCVMAR} and (d) DukeMTMC-VideoReID dataset~\cite{WuYu2018CVPROneShotforVideoReID}. In each dataset, we sample two video sequences from one person, and the video sequences are captured by disjoint cameras. For the sake of privacy, we use a black region to cover the face in each frame.}}\label{fig_samp}
\end{figure}

\begin{figure}[!ht]
\centering
\subfigure[Samples from the CUHK01 dataset.]{
\includegraphics[width=5.4cm]{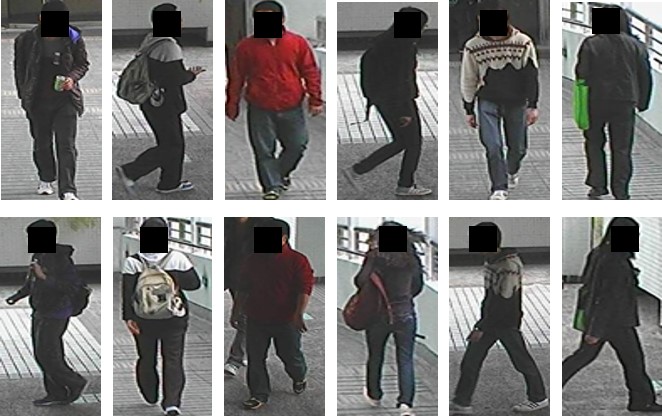}
\label{samp_cuhk}
}
\quad
\subfigure[Samples from the DukeMTMC-reID dataset.]{
\includegraphics[width=5.4cm]{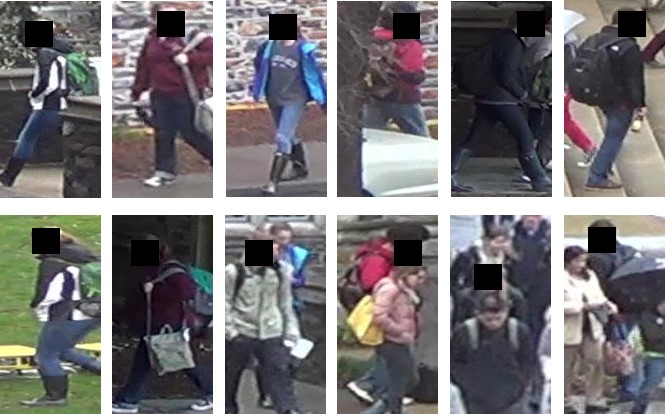}
\label{samp_iduke}
}
\caption{\small{Samples from: (a) CUHK01~\cite{LiWei2012ACCVCUHK01}, and (b) DukeMTMC-reID~\cite{ristani2016MTMC}. In each dataset, we sample two images from each person, and the video sequences are captured by disjoint cameras. For the sake of privacy, we use a black region to cover the face in each frame.}}\label{fig_samp_i}
\end{figure}

In the main paper, we have evaluated our attention mechanism across four video person re-ID datasets and two image person re-ID datasets. Here, we show some samples from the aforementioned datasets, in Fig.~\ref{fig_samp} and Fig.~\ref{fig_samp_i}. In each pedestrian bounding box, we use a black region to cover the face parts for the sake of privacy.

\section{Ablation Study for the Baseline Network}
In this section, extensive experiments are performed to choose a proper setting for the baseline network, including the number frames to use from a video clip, dimensionality of the video feature embedding and the training strategies (\eg, pre-training and random erasing~\cite{zhong2017random}). This ablation studies are performed on the iLIDS-VID~\cite{WangTaiqing2016TPAMIiiLIDS-VID} and the MARS~\cite{ZhengLiang2014ECCVMAR} datasets.  

\noindent\textbf{Number of Frames in Video Clip.}
First, we perform experiments with a different number of frames (\ie, $t$) in a video clip. When $t = 1$, it is reduced to the single image-based model. From Table~\ref{table:Frames}, we observe that $t = 4$ achieves the highest accuracy in both R-1 and mAP values. Thus we use $t = 4$ in our work.

\begin{table}[!ht]
\caption{\footnotesize{Effect of the number of frames in a video clip on the iLIDS-VID~\cite{WangTaiqing2016TPAMIiiLIDS-VID} and the MARS~\cite{ZhengLiang2014ECCVMAR} datasets.}}
\begin{center}
\scalebox{1}{
\begin{tabular}{c|c|c|c|c|c}
\toprule[2pt]
\hline
\multicolumn{2}{c|}{}  & \multicolumn{2}{c|}{~~iLIDS-VID~~} & \multicolumn{2}{c}{~~MARS~~} \\
\hline
\multicolumn{2}{c|}{~~Num of Frames~~}&~~R-1~~&~~mAP~~&~~R-1~~&~~mAP~~\\
\hline
(\romannumeral1) &$t = 1$ &76.3&84.2&79.2&74.3\\
(\romannumeral2) &$t = 2$ &79.3&86.1&81.5&75.6\\
(\romannumeral3) &$t = 4$ &\textbf{80.0}&\textbf{87.1}&\textbf{82.3}&\textbf{76.2}\\
(\romannumeral4) &$t = 8$ &79.6&86.4&82.1&76.0\\
\hline
\hline
\toprule[2pt]
\end{tabular}
}
\end{center}\label{table:Frames}
\end{table}

\noindent\textbf{Dimensionality of Video Feature Embedding.}
The dimension, \ie, $\mathrm{D}_v$, of the video feature embedding is evaluated and illustrated in Table~\ref{table:Dim} on both the iLIDS-VID~\cite{WangTaiqing2016TPAMIiiLIDS-VID} and the MARS~\cite{ZhengLiang2014ECCVMAR} datasets. On iLIDS-VID, it is clear that the video feature embedding with $\mathrm{D}_v = 1024$ performs better for both R-1 and mAP accuracy. Therefore, we choose $\mathrm{D}_v = 1024$ as the dimension of the feature embedding across all datasets. On the MARS dataset, we observe that R-1 has the peak value when $\mathrm{D}_v = 512$, while mAP achieves the peak value when $\mathrm{D}_v = 1024$. However, the mAP value in $\mathrm{D}_v = 512$ is much lower than that in $\mathrm{D}_v = 1024$. Thus we also choose $\mathrm{D}_v = 1024$ for MARS.

\begin{table}[!ht]
\caption{\footnotesize{Effect of the dimensionality of video feature embedding on the iLIDS-VID~\cite{WangTaiqing2016TPAMIiiLIDS-VID} and the MARS~\cite{ZhengLiang2014ECCVMAR} datasets.}}
\begin{center}
\scalebox{1}{
\begin{tabular}{c|c|c|c|c|c}
\toprule[2pt]
\hline
\multicolumn{2}{c|}{}  & \multicolumn{2}{c|}{~~iLIDS-VID~~} & \multicolumn{2}{c}{~~MARS~~} \\
\hline
\multicolumn{2}{c|}{~Dim of Embedding~~}&~~R-1~~&~~mAP~~&~~R-1~~&~~mAP~~\\
\hline
(\romannumeral1) &$\mathrm{D}_v = 128$ &72.0&81.0&82.0&75.1\\
(\romannumeral2) &$\mathrm{D}_v = 256$ &73.3&82.5&82.4&76.3\\
(\romannumeral3) &$\mathrm{D}_v = 512$ &76.6&85.5&\textbf{82.6}&75.2\\
(\romannumeral4) &$\mathrm{D}_v = 1024$ &\textbf{80.0} &\textbf{87.1}&82.3&\textbf{76.2}\\
(\romannumeral5) &$\mathrm{D}_v = 2048$ &79.6&86.5&82.0&75.6\\
\hline
\hline
\toprule[2pt]
\end{tabular}
}
\end{center}\label{table:Dim}
\end{table}

\noindent\textbf{Training Strategies.}
We further analyze the effect of different training strategies of the deep network (\eg, random erasing, pre-training model) in Table \ref{table:TrainingComp} on both the iLIDS-VID and the MARS datasets. Here, $f_{\theta}$ denotes the backbone network (see Fig. \textcolor{red}{3} in the main paper). Pre-T and RE denote pre-training on imageNet \cite{russakovsky2015imagenet} and random erasing data augmentation, respectively. This table reveals that both training components of pre-training (\ie, Num (\romannumeral2)) and random erasing (\ie, Num (\romannumeral3)) improve the R-1 and mAP values, compared to the baseline (\ie, Num (\romannumeral1)). In addition, the network continues to improve its performance when both training strategies are employed, showing that those two training strategies work in a complementary fashion. Thus we choose the network with the pre-trained model and random erasing as our baseline network.

\begin{table}[!ht]
\caption{\footnotesize{Effect of the different training strategies on the iLIDS-VID~\cite{WangTaiqing2016TPAMIiiLIDS-VID} and the MARS~\cite{ZhengLiang2014ECCVMAR} datasets. $f_{\theta}$, Pre-T and RE denote backbone network, pre-training and random erasing, respectively}}
\begin{center}
\scalebox{1}{
\begin{tabular}{c|c|c|c|c|c}
\toprule[2pt]
\hline
\multicolumn{2}{c|}{}  & \multicolumn{2}{c|}{~~iLIDS-VID~~} & \multicolumn{2}{c}{~~MARS~~} \\
\hline
\multicolumn{2}{c|}{~~Model~~}&~~R-1~~&~~mAP~~&~~R-1~~&~~mAP~~\\
\hline
(\romannumeral1) &$f_{\theta}$   &60.8&67.6&76.4&71.8\\
(\romannumeral2) &$f_{\theta}$ + Pre-T   &70.8&81.6&81.1&75.4\\
(\romannumeral3) &$f_{\theta}$ + RE   &65.3&74.6&78.8&74.5\\
(\romannumeral4) &~~~$f_{\theta}$ + Pre-T + RE~~~&\textbf{80.0} &\textbf{87.1}&\textbf{82.3}&\textbf{76.2}\\
\hline
\hline
\toprule[2pt]
\end{tabular}
}
\end{center}\label{table:TrainingComp}
\end{table}

\section{Ablation Study for Set Aggregation Cell}
In this section, we perform experiments for parameter selection in the set aggregation cell.

\noindent\textbf{Effectiveness of Dimension Reduction in the Self-gating Layers.} The dimension of the hidden layer in the self-gating layers (\ie, $\varpi$ and $\psi$) of the set aggregation block is studied. The dimension of the hidden layer is reduced by a factor of $r$ (\ie, $\mathrm{D}_{\mathrm{hid}} = 2048 / r$). Table~\ref{table:reduction_sa} reveals that setting $r = 16$ achieves good performance in both datasets; thus we use this value for the set aggregation cell.

\begin{table}[!ht]
\caption{\footnotesize{Effect of dimension reduction in the self-gating layers on the iLIDS-VID~\cite{WangTaiqing2016TPAMIiiLIDS-VID} and the MARS~\cite{ZhengLiang2014ECCVMAR} datasets.}}
\begin{center}

\scalebox{1}{
\begin{tabular}{c|c|c|c|c|c}
\toprule[2pt]
\hline
\multicolumn{2}{c|}{}  & \multicolumn{2}{c|}{~~iLIDS-VID~~} & \multicolumn{2}{c}{~~MARS~~}\\
\hline
\multicolumn{2}{c|}{~~Reduction Ratio~~}&~~R-1~~&~~mAP~~&~~R-1~~&~~mAP~~\\
\hline
(\romannumeral1) &~~Only CRA~~&87.0 &90.6 & 86.8 & 81.6\\
\hline
(\romannumeral2) &$r = 2$    &88.2&91.2&87.2&82.1\\
(\romannumeral3) &$r = 4$    &88.4&91.6&87.6&82.4\\
(\romannumeral4) &$r = 8$    &88.5&91.7&\textbf{87.9}&82.8\\
(\romannumeral5) &$r = 16$   &\textbf{88.7}&\textbf{91.9
}&\textbf{87.9}&\textbf{83.0}\\
(\romannumeral6) &$r = 32$   &87.9&90.9&87.6&82.2\\
\hline
\hline
\toprule[2pt]
\end{tabular}
}
\end{center}\label{table:reduction_sa}
\end{table}

\section{Visualization}
\begin{figure}[!ht] 
\centering
\includegraphics[width = 0.96\linewidth]{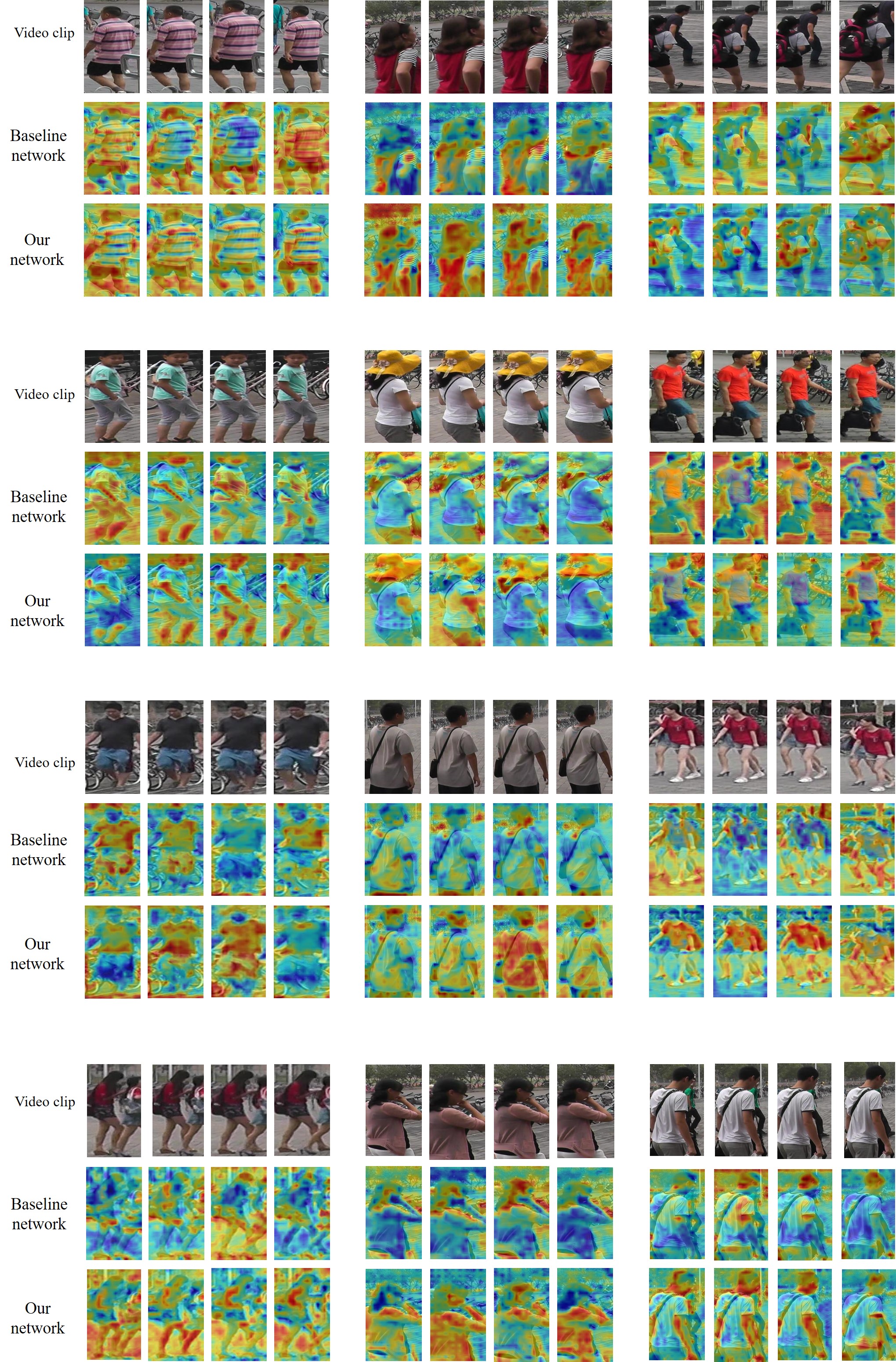}
\caption{\footnotesize{Visualization of feature maps. We sample video clips from different pedestrians and visualize the feature maps.}} \label{fig:Vis_supp}
\end{figure}

Fig.~\ref{fig:Vis_supp} shows additional visualizations of feature maps for qualitative study. In the visualizations, we can clearly observe that our attention module has the capacity to focus more on the foreground areas and ignore some background areas, which boosts the baseline network to achieve the state-of-the-art performance on the video pedestrian task.

\end{document}